\documentclass[manuscript,authorversion,screen,nonacm]{acmart}
\AtBeginDocument{%
  }

\setcopyright{acmlicensed}
\copyrightyear{2024}
\acmYear{2024}
\acmDOI{XXXXXXX.XXXXXXX}

\acmJournal{TiiS}
\acmVolume{14}
\acmNumber{3}
\acmArticle{1}
\acmMonth{12}

\usepackage[ruled,vlined,linesnumbered,noresetcount]{algorithm2e}
\usepackage{algorithmic}
\usepackage{subcaption}
\usepackage{tabularx}
\usepackage{booktabs}
\usepackage{enumitem}
\usepackage{todonotes}
\usepackage{lipsum}

\newcommand{\methodName}{HAIFAI}

\begin{teaserfigure}
    \centering
    \includegraphics[width=0.80\textwidth]{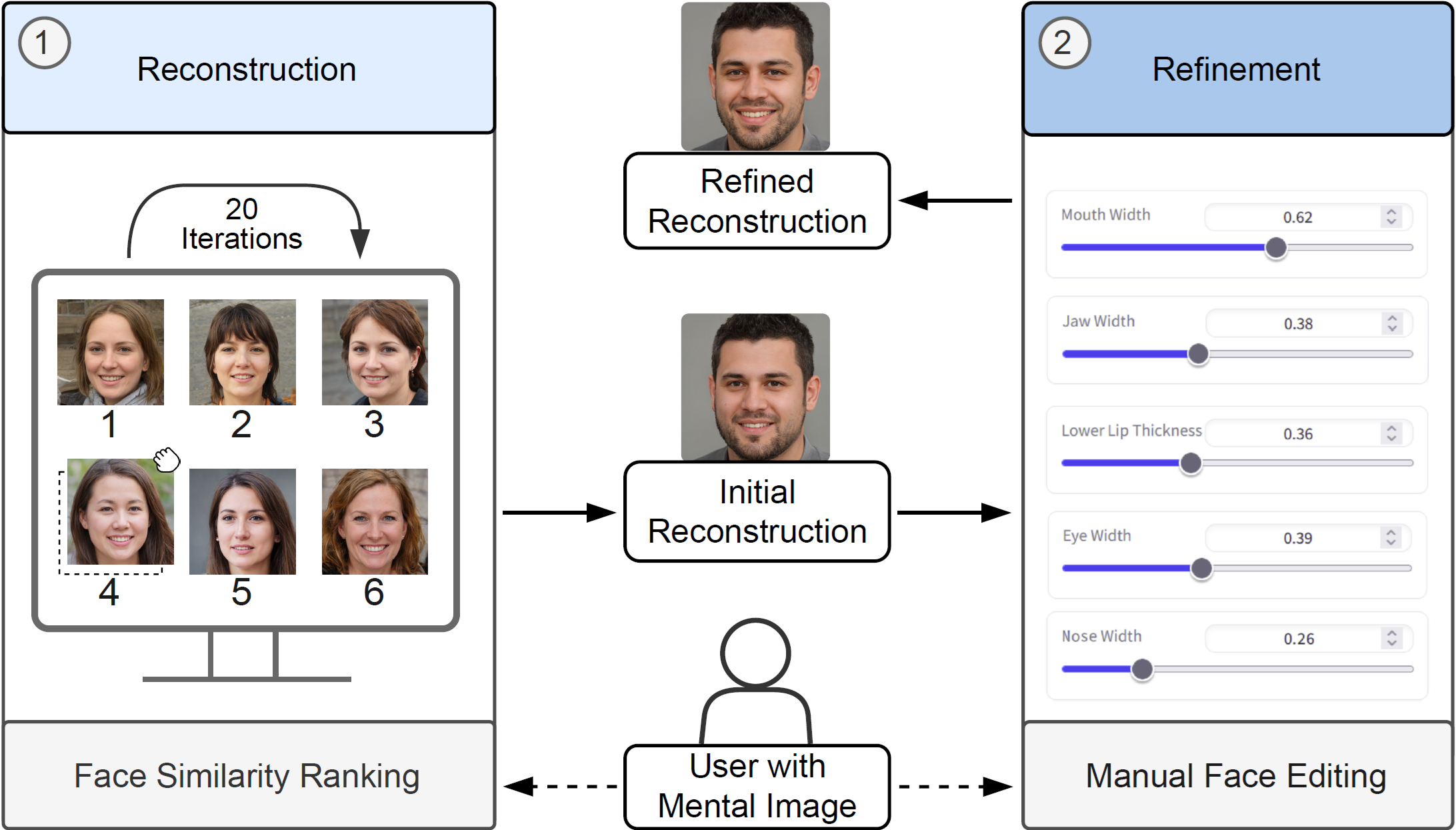}
    \caption{\methodName\ is an interactive human-AI system designed to reconstruct face images from a user's mental image visually. In this system, users iteratively rank auxiliary face images -- random face images of the same age group and sex -- based on their perceived visual similarity to the mental image they hold in mind. This feedback is utilised to extract relevant image features, which are then combined by the AI system to reconstruct the user's mental image. Following the initial reconstruction, users can adjust facial features further using a slider interface to refine the visual reconstruction.
    }
    \label{fig:teaser}
\end{teaserfigure}

\begin{document}

\title{\methodName: \underline{H}uman-\underline{AI} Interaction for Mental \underline{Fa}ce Reconstruct\underline{i}on}
%% 

\author{Florian Strohm}
\email{florian.strohm@vis.uni-stuttgart.de}
\orcid{0000-0002-3787-3062}
\affiliation{%
  \institution{University of Stuttgart}
  \country{Germany}
}

\author{Mihai Bâce}
\orcid{0000-0002-1446-379X}
\affiliation{%
  \institution{KU Leuven}
  \country{Belgium}
}
\authornote{Part of this work was conducted while at the University of Stuttgart.}
\email{mihai.bace@kuleuven.be}

\author{Andreas Bulling}
\orcid{0000-0001-6317-7303}
\affiliation{%
  \institution{University of Stuttgart}
  \country{Germany}
}
\email{andreas.bulling@vis.uni-stuttgart.de}

\renewcommand{\shortauthors}{Florian Strohm, Mihai Bâce, \& Andreas Bulling}


\begin{abstract}
We present \methodName\ -- a novel two-stage system where humans and AI interact to tackle the challenging task of reconstructing a visual representation of a face that exists only in a person's mind. 
In the first stage, users iteratively rank images our reconstruction system presents based on their resemblance to a mental image.
These rankings, in turn, allow the system to extract relevant image features, fuse them into a unified feature vector, and use a generative model to produce an initial reconstruction of the mental image.
The second stage leverages an existing face editing method, allowing users to manually refine and further improve this reconstruction using an easy-to-use slider interface for face shape manipulation.
To avoid the need for tedious human data collection for training the reconstruction system, we introduce a computational user model of human ranking behaviour.
For this, we collected a small face ranking dataset through an online crowd-sourcing study containing data from 275 participants. 
We evaluate \methodName\ and an ablated version in a 12-participant user study and demonstrate that our approach outperforms the previous state of the art regarding reconstruction quality, usability, perceived workload, and reconstruction speed.
%\methodNameS\ achieves even better reconstruction quality at the cost of reduced usability, perceived workload, and increased reconstruction time.
We further validate the reconstructions in a subsequent face ranking study with 18 participants and show that \methodName\ achieves a new state-of-the-art identification rate of $60.6\%$. 
These findings represent a significant advancement towards developing new interactive intelligent systems capable of reliably and effortlessly reconstructing a user's mental image.
\end{abstract}




\maketitle

\section{Introduction}

Many humans are visual thinkers~\cite{arnheim2023visual}, i.e., they heavily rely on mental imagery in their everyday life -- visual representations of objects, faces, or concepts only available in people's minds. 
For visual thinkers, the ability to ``see'' scenarios, memories, or future possibilities is critical for how they process information, solve problems, or make decisions~\cite{kim2021roles,sims2019perceptual,jeannerod1995mental,moulton2009imagining,pearson2015mental}.
The prospect of visually reconstructing such mental images using computational methods has long fascinated researchers and has led to substantial research efforts in this area.
Mental image reconstruction not only promises to improve our understanding of the human visual system by analysing how visual information is stored in the brain.
Artificial intelligent (AI) systems that can understand human mental processes also hold significant promise for enhancing human-AI interaction.

Mental image reconstruction as a computational task is profoundly challenging given the complex neural encoding of mental images in the human brain \citep{naselaris2015voxel}. 
Prior works have used brain sensing techniques, such as electroencephalography (EEG) \citep{zheng2020decoding, shatek2019decoding, bruera2022exploring, date2019deep} or functional magnetic resonance imaging (fMRI) \citep{guccluturk2017reconstructing,dado2022hyperrealistic,beliy2019voxels,shen2019deep,lin2019dcnn,vanrullen2019reconstructing,seeliger2018generative, scotti2024reconstructing, ozcelik2023natural, takagi2023high}. 
These methods, though promising, are limited by their invasive nature (EEG) or impractical for everyday use due to their cost and technical complexity (fMRI). 
Consequently, recent works have instead explored passive monitoring of human gaze for mental image reconstruction. 
Although gaze has been shown to reflect cognitive processes, such as visual memory \cite{bulling11_ubicomp}, and is thus often referred to as a ``window into the mind'', previous works have achieved only limited success in terms of reconstruction quality \citep{sattar2017predicting,sattar2020deep,strohm2021neural,strohm2022facial}.

We propose a more practical approach to mental image reconstruction where a human user and an artificial intelligence (AI) system interactively work together to reconstruct mental images using \textit{active} user feedback. 
While our approach can, by design, be used with all mental images, we specifically focus on human faces given the importance of face perception, e.g., in social interactions, and given highly relevant practical applications, such as reconstructing a suspect's face from a witness's memory in criminology.
Existing methods for facial composite generation can be categorised into constructive, holistic, and hybrid. 
Constructive approaches offer extensive catalogues of facial features for users to choose from, such as different eye, nose, and mouth shapes and appearances \citep{ellis1978critical,Laughery1980,christie1981photofit,koehn1997constructing}. 
The main drawback of constructive approaches is that humans struggle to identify individual features accurately in isolation and instead seem to recall faces holistically \citep{farah1998special}. 
Consequently, holistic methods, such as EvoFIT \citep{frowd2004evofit}, allow users to assemble entire faces iteratively through evolutionary algorithms. 
However, guiding a holistic generation can be more challenging than selecting specific features from a catalogue. 
Hybrid methods (e.g., CG-GAN~\citep{zaltron2020cg}) combine both approaches' advantages by allowing interactive full-face refinement while maintaining control over specific facial features. 
However, current hybrid approaches are limited by slow exploration of the high-dimensional face appearance space and in terms of user control.

To address these challenges, we introduce \methodName\ -- an interactive mental face reconstruction system designed to maximise the utility of user feedback without depending on random exploration. 
Our method involves users iteratively ranking sets of face images based on their resemblance to their mental image. 
This approach significantly simplifies the users' tasks compared to prior methods, such as CG-GAN, which require users to combine various mechanisms to reconstruct their mental images.
Our system then extracts facial appearance information from these rankings over multiple iterations. 
This information is integrated using an end-to-end, data-driven model that predicts a feature vector that encodes likely facial features. 
Rather than iteratively searching the face space as done in evolutionary-based algorithms, our system holistically integrates user feedback, using the available information's full potential.
To visually decode the mental image, \methodName\ uses a state-of-the-art generative model capable of generating realistic face images\citep{karras2020analyzing}. 
Due to the impracticality of collecting large amounts of human ranking feedback for training \methodName, we propose a computational user model to simulate this process. 
This model uses a pre-trained face identification network \citep{deng2019arcface}, fine-tuned on a face similarity task with human labels crowd-sourced from 275 participants via Amazon Mechanical Turk (AMT). 
This enables the generation of synthetic human rankings to train our system.
In the second stage, we use UP-FacE~\cite{strohm2024upface} to allow users to further refine the initial reconstruction with an easy-to-use slider interface.
Providing an initial reconstruction based on user rankings that closely align with their mental image facilitates easier and more efficient subsequent manual editing of the face.

\noindent
Our work makes the following contributions:
\begin{enumerate}[leftmargin=0.5cm,itemsep=0.25cm,topsep=0pt]
    \item We propose \methodName, a two-stage interactive human-AI mental face image reconstruction system. First, our novel method integrates explicit user ranking and reconstructs an initial image. Second, UP-FacE enables fine-tuned control over the facial features of the face reconstruction.
    \item We introduce a computational user model for ranking, trained on a novel crowd-sourced dataset of human face ranking information. This allows us to generate the data required for training the first stage of \methodName.
    \item We demonstrate the effectiveness of \methodName\ through evaluations in two user studies, highlighting improvements in usability, reconstruction speed, and quality over state-of-the-art methods. Additionally, our system achieves a new state-of-the-art performance for deep learning-based techniques in human identification rates, which is particularly significant for critical applications such as forensics.
\end{enumerate}

\section{Related Work}

Our work is related to previous works on mental face reconstruction using 1) implicit and 2) explicit user feedback, as well as 3) face editing. 

\subsection{Mental Face Reconstruction with Implicit User Feedback}

Prior research on reconstructing faces using implicit user feedback mechanisms has predominantly relied on EEG, fMRI, or gaze data. 
Pioneering work in fMRI-based face reconstruction by \citet{cowen2014neural} involved mapping fMRI signals to principal components of facial structures, also known as Eigenfaces, and reconstructing faces through linear combinations of these components. 
\citet{nestor2016feature} have developed a method to infer a facial feature space directly from fMRI data. 
They employed an SVM to distinguish face identities based on fMRI responses and created template faces for each axis within the derived feature space.
These templates were then used to reconstruct faces from fMRI signals through interpolation. 
This methodology was later adapted by \citet{nemrodov2018neural} for EEG-based mental face reconstruction.
A different approach was introduced by \citet{vanrullen2019reconstructing}, who used a VAE-GAN architecture to encode facial images into a low-dimensional latent space. 
They then mapped fMRI responses to this latent space using a simple regression model, allowing for face reconstruction via the VAE-GAN decoder.
Remarkably, they could reconstruct faces from fMRI responses even when participants merely thought of a face without visual stimuli. 
Building on this, \citet{dado2022hyperrealistic} have recorded fMRI responses to synthetic faces produced by a pre-trained PGGAN~\cite{karras2017progressive}. 
Similar to \citet{vanrullen2019reconstructing}, they used linear regression to map fMRI data to the GAN's latent space for face reconstruction.
Recent works have used diffusion models as powerful image generation backbones, improving image reconstruction quality. This is accomplished either by mapping the fMRI response onto the latent space of the model~\cite{takagi2023high, scotti2024reconstructing} or by refining an initial reconstruction from a variational autoencoder~\cite{ozcelik2023natural}.
While all of these fMRI and EEG-based methods have demonstrated potential, they are also all limited by their invasiveness, high costs, and the challenge of generalising models to unseen users as the relevant brain regions differ in size and neural activity.

To address the limitations of these invasive techniques, \citet{strohm2021neural} have introduced a gaze-based method for mental face reconstruction.
In that study, participants were asked to look at various faces and their gaze patterns were analysed to predict relevant facial features.
These features were then combined using a pre-trained decoder to generate mental images. 
Although less invasive and costly, this first method required prior knowledge of the target face, which is prohibitive for most practical applications.
They later developed an improved approach that eliminated this requirement but still relied on a controlled environment of human-like faces and accurate gaze data ~\cite{strohm2022facial}. 
Given the challenges of mental face reconstruction using implicit feedback, we propose methods that use explicit feedback instead.
Our methods operate in a less constrained domain of real faces and significantly improve reconstruction quality and, thus, practical usefulness.

\subsection{Mental Face Reconstruction with Explicit User Feedback}

A large body of work has explored the reconstruction of mental images through the use of explicit feedback mechanisms, such as user selection, ranking, and manual editing.
Early systems for face creation were based on the constructive paradigm, allowing users to select individual facial features from extensive template catalogues~\cite{ellis1978critical, Laughery1980, christie1981photofit, koehn1997constructing}. 
However, such methods were constrained by the holistic nature of human face perception, as individuals struggle to identify isolated features accurately~\cite{farah1998special}.
To address this limitation, \citet{frowd2004evofit} have introduced the popular EvoFIT system that allows for holistic face interpolation within the Eigenface space:
Users iteratively select faces that resemble their mental image, and the system generates new faces based on the selected Eigenfaces using an evolutionary algorithm. 
A similar approach has been proposed by \citet{gibson2009new} with added functionality for adjusting for age and facial details, such as wrinkles. 
Further advances in this field included the deep interactive evolution method by \citet{bontrager2018deep}, which applied an evolutionary algorithm in the latent space of a pre-trained GAN to enhance image quality, and \citet{xu2019generating} used a GAN conditioned on facial landmarks and iteratively refined the landmarks through user feedback to improve reconstruction quality. 
\citet{zaltron2020cg} introduced CG-GAN, which integrates a holistic evolutionary algorithm with constructive functionalities. 
This approach allows users to modify faces along identified axes within the GAN's latent space using binary face labels (e.g., \textit{glasses}, \textit{beard}).
Lastly, \citet{chiu2020human} have proposed a method to explore one-dimensional subspaces of a pre-trained GAN and modify faces using sliders until the desired face was obtained.
Our method fundamentally differs in that it learns to interpret user feedback through an end-to-end training process, bypassing the need for traditional evolutionary algorithms or random exploration strategies.
This approach enables our system to holistically integrate user feedback, leveraging the contained information more effectively. 
As a result, we achieved improved reconstructed image quality and reduced reconstruction times.

\subsection{Digital Face Editing}

Utilising prior work in digital face editing, the second stage of \methodName\ allows users to perform additional manual face edits after the initial reconstruction was obtained.
%For this we require a face editing method, which can be grouped into different categories.
Digital face editing is a challenging but well-established task in computer vision that has received increasing attention, particularly in recent years.
Related methods have typically used generative models that can produce high-quality facial images from a latent vector encoding facial features.
Face editing is performed by manipulating these corresponding latent vectors.
There are several distinct classes of methods, including unsupervised methods, mask-based methods, text-based methods, 3D-based methods, and attribute-based approaches.
Unsupervised face editing methods decompose generative models' latent space or weights to uncover semantic editing dimensions. 
GANSpace~\cite{harkonen2020ganspace} uses PCA on the latent space, while Shen et al.~\cite{shen2021closed} have decomposed generator weights, and Niu et al.~\cite{niu2023disentangling} have refined semantic directions. 
Although these methods do not require labelled data, they are limited in their ability to uncover meaningful editing dimensions in the latent space and often exhibit higher entanglement with other features. 
This entanglement can lead to unwanted modifications in other facial features.
DragGAN~\cite{pan2023drag} allows manual adjustment of facial landmarks, demanding significant user effort and unsuitability for automation.
Mask-based techniques condition generative models on face masks for control, such as sketch-based inpainting~\cite{portenier2018faceshop, chen2020deepfacedrawing, chen2021deepfaceediting} or mask manipulation edits~\cite{gu2019mask, song2019geometry, lee2020maskgan, ling2021editgan, sun2022ide, sun2022fenerf}. 
These methods, however, require skill and effort to modify the segmentation masks properly.
Text-based face editing methods combine mask techniques with text-guided edits~\cite{xia2021tedigan, huang2023collaborative}, such as by integrating generators with the CLIP encoder~\cite{patashnik2021styleclip} or by allowing localised edits and broader text prompts~\cite{hou2022feat, sun2022anyface}. 
While text-based methods offer a more user-friendly interface and high-level control over many aspects of a face, they lack fine-grained editing precision.
3D-based methods translate a 3D face model to a real image~\cite{tewari2020stylerig,deng2020disentangled,tewari2020pie,medin2022most,kowalski2020config} and are great for novel-view synthesis, lighting manipulation or transferring expressions.
However, semantic face shape editing necessitates significant 3D modelling efforts.
Finally, early attribute-based methods have offered binary control over predefined face attributes, but this approach often resulted in unintended changes~\cite{choi2018stargan, xiao2018elegant, lu2018attribute, zhang2018generative, guo2019mulgan, yang2021l2m}. 
Later methods have used attention mechanisms and classifiers for localised edits~\cite{zhang2018generative, kwak2020cafe, wei2020maggan, he2020pa, he2019attgan, sun2022pattgan, gao2021high} or SVMs for smooth manipulation~\cite{shen2020interfacegan, han2021disentangled}. 
Other research has focused on discovering directions in a generator's latent space for plausible edits~\cite{yao2021latent, hou2022guidedstyle, khodadadeh2022latent, abdal2021styleflow, wu2021stylespace, yang2021discovering}. 
\citet{strohm2024upface} have recently proposed UP-FacE -- a method that uses landmark annotations to find such editing directions in latent space.
UP-FacE offers user-predictable and fine-grained control over various shape-based facial features, including the eyebrows, eyes, nose, and mouth. These features have been demonstrated to be crucial for face recognition~\cite{sinha2006face}.
\section{Interactive Mental Face Reconstruction}
\label{sec:method}

\begin{figure*}[ht]
    \centering
    \includegraphics[width=.7\textwidth]{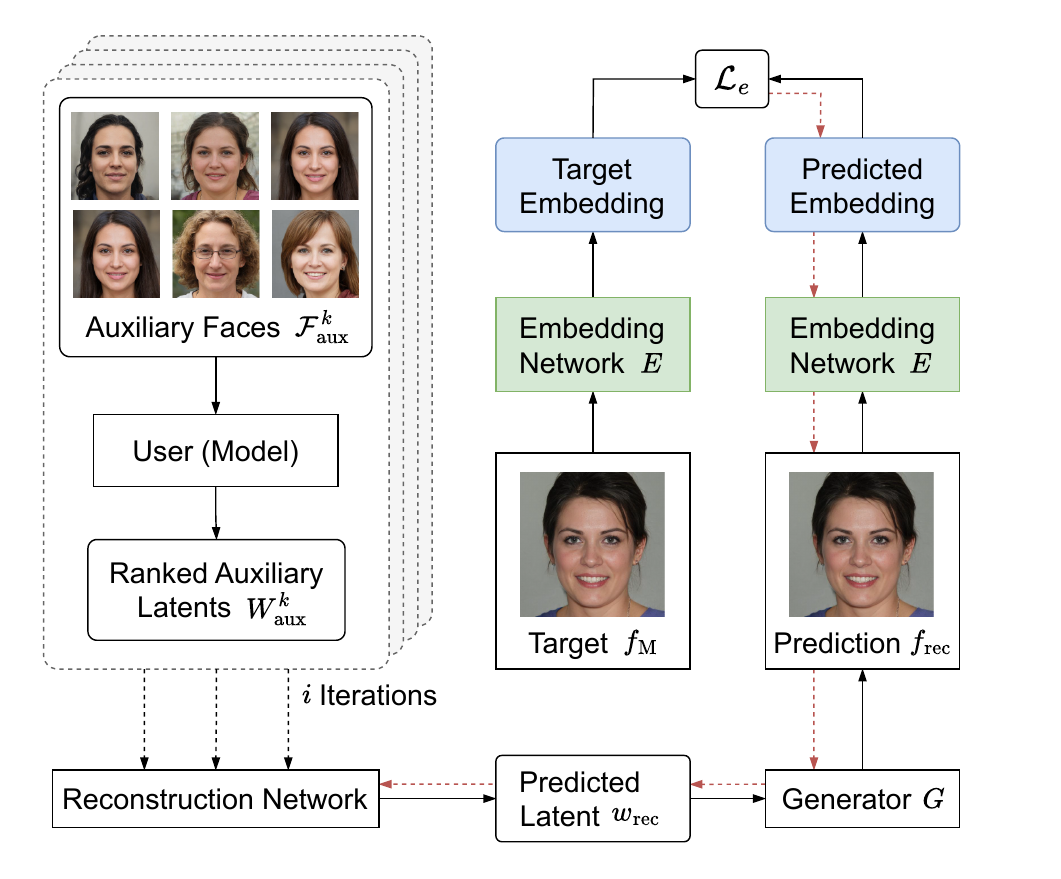}
    \caption{Our system \methodName\ first shows users sets of auxiliary faces over multiple rounds, asking them to rank these faces based on their resemblance to their mental image. A reconstruction network then predicts the latent vector corresponding to this mental image using the ranked auxiliary latents. A pre-trained generator decodes this vector to recreate the image. The target and reconstructed face images are then passed through a pre-trained embedding network to optimise our reconstruction network based on the similarity of their embeddings. The dotted red arrows indicate the gradient path to train our reconstruction network.
    }
    \label{fig:MFRS_architecture}
\end{figure*}

The objective of mental face reconstruction is to create a visual representation, $f_\text{rec}$, of a mental face image $f_\text{m}$ only present in a person's mind. 
The goal is to generate $f_\text{rec}$ such that it resembles the same person identity (id) as best as possible: $f_\text{rec} \stackrel{id}{=} f_\text{m}.$
Our approach for mental face reconstruction involves a two-step process: First, a human user and our system collaborate to iteratively generate a face reconstruction based on the user's explicit ranking feedback. 
In a second step, if desired, the user can easily fine-tune the visual reconstruction using UP-FacE~\cite{strohm2024upface} by adjusting various semantic facial features, such as nose width and lip size.

\autoref{fig:MFRS_architecture} presents an overview of the first step in our approach.
The high-level concept involves utilising a generative model $G$ for faces. 
This model accepts a latent vector $w$ as input, which encodes facial features and subsequently generates a corresponding face image $f$. 
The primary objective of our system is to predict a latent vector $w_\text{rec}$ that encodes the relevant features of a user's mental image. 
This allows the generative model $G$ to accurately produce a visual representation $f_\text{rec}$ of the mental image.
To generate the image $f_\text{rec}$ from the latent vector $w_\text{rec}$ we use a pre-trained StyleGAN2~\cite{karras2020analyzing} model -- a generative model trained on the FFHQ~\cite{karras2019style} faces dataset.
Importantly, StyleGAN2 first maps the input to a disentangled latent space $\mathcal{W}$, which has a clear advantage over the original input latent space $\mathcal{Z}$:
The $\mathcal{W}$ space was shown to exhibit less entanglement~\cite{karras2020analyzing, khodadadeh2022latent, abdal2021styleflow, wu2021stylespace}, resulting in better controllability over the generation process.
This disentanglement means that changes to specific latent dimensions in $\mathcal{W}$ tend to correspond to semantically meaningful edits to the generated image, such as modifying expressions, hairstyles, or other facial attributes, without inadvertently affecting unrelated features.
To generate the initial reconstruction, our system presents the user a selection of $n$ predefined auxiliary face images, denoted as $\mathcal{F}_\text{aux} = {f_1,...,f_n}$, which were generated randomly.
The user is asked to rank these auxiliary images based on their resemblance to the mental face image $f_\text{m}$ they have in mind. 
\methodName\ uses these rankings to determine the optimal latent vector $w_\text{rec}$ that, when passed through StyleGAN2's generator $G$, results in an image $f_\text{rec}$ that matches the mental face image $f_\text{m}$:
\[
G(w_\text{rec}) = f_\text{rec} \stackrel{id}{=} f_\text{m}.
\]

The user's ability to accurately rank $n$ images diminishes as the number of auxiliary images increases due to the factorial increase in potential rankings.
To address this, we adopt an iterative approach, limiting the number of images per iteration to six. 
This simplifies the user's task and results in less noisy rankings.
In each iteration $i$, the system presents a different set of six auxiliary images $\mathcal{F}_\text{aux}^i = {f_1^i,...,f_6^i}$, generated using StyleGAN2 by randomly sampling latent vectors $W_\text{aux}^i = {w_1^i,...,w_6^i}$. 

Given the challenge of comparing and ranking faces across different age and sex groups, our system categorises faces based on sex (female or male) and age (above or below 40 years). 
This approach is commonly used in mental face reconstruction systems~\cite{zaltron2020cg}.
We employed the InsightFace\footnote{\url{https://insightface.ai/projects}} toolbox to automatically label the sex and age of faces randomly generated using StyleGAN2. 
This labelling process allowed us to create distinct sets of auxiliary images for each sex and age group.
\methodName\ receives the sex and age information as input and uses the corresponding sets of auxiliary images to proceed with the reconstruction process.
In the following, we describe each of these components in detail.

\begin{figure*}[t]
    \centering
    \includegraphics[width=0.6\textwidth]{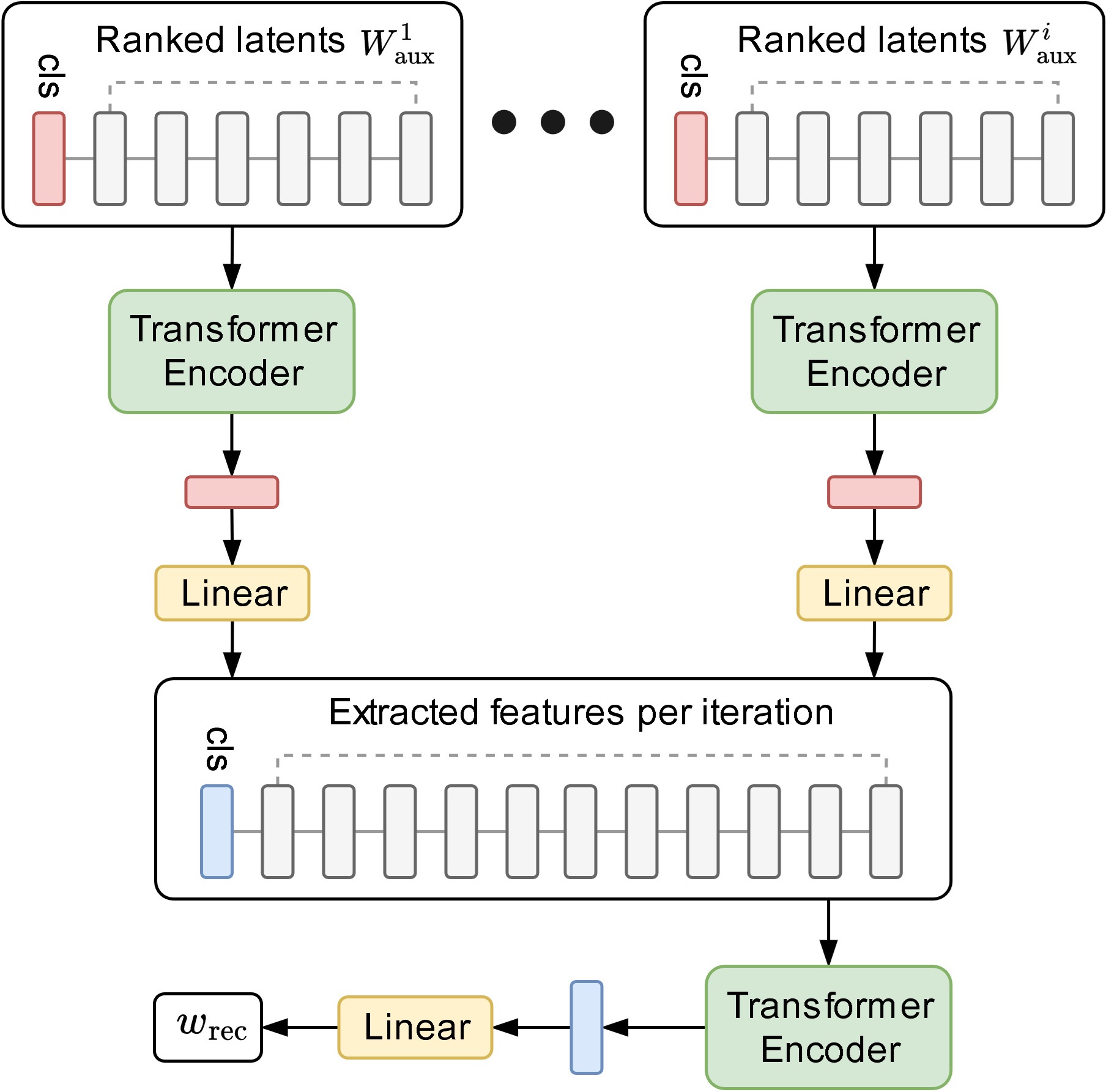}
    \caption{Architecture of our reconstruction network. The reconstruction network processes a tuple of six auxiliary latent vectors ordered by user rankings for each iteration. For each tuple, a learnable 512-dimensional token (cls) is appended. These tuples are then fed into a Siamese transformer encoder, and the cls tokens are extracted at the end and passed through a linear layer. The resulting feature vectors are combined with another learnable token and processed by another transformer encoder. The class token is extracted again at the output and passed through a final linear layer to produce the reconstructed latent vector of the mental image.
    }
    \label{fig:reconstruction_network}
\end{figure*}

\subsection{Reconstruction Network}

The objective of the reconstruction network is to predict a latent vector $w_\text{rec}$ such that this vector, when processed by the generator $G$, produces a face image closely resembling the person identity of the mental image $f_m$. 
The architecture of the reconstruction network is shown in \autoref{fig:reconstruction_network}.
The reconstruction network takes as input multiple tuples $W_\text{aux}^i$, each containing six auxiliary latent vectors. 
These vectors within each tuple are ordered according to the user's rankings.
For each tuple of latent vectors, we append an additional learnable embedding token of size $512$, commonly referred to as the \textit{class token (cls)}~\cite{devlin2018bert} and add positional encodings such that the model can extract features based on the ranking of the latent vectors.
Each tuple is then processed by a Siamese transformer encoder~\cite{vaswani2017attention}, the class token is read out at the end and finally passed through a linear layer, resulting in a $512$ dimensional vector containing relevant features for the respective iteration.
The feature vectors extracted from each iteration are stacked with another learnable token and passed through a second transformer encoder model.
The class token is again extracted at the output of this transformer and passed through a final linear layer, resulting in the reconstructed latent vector $w_\text{rec}$.

\begin{figure}[t]
    \centering
    \includegraphics[width=0.7\linewidth]{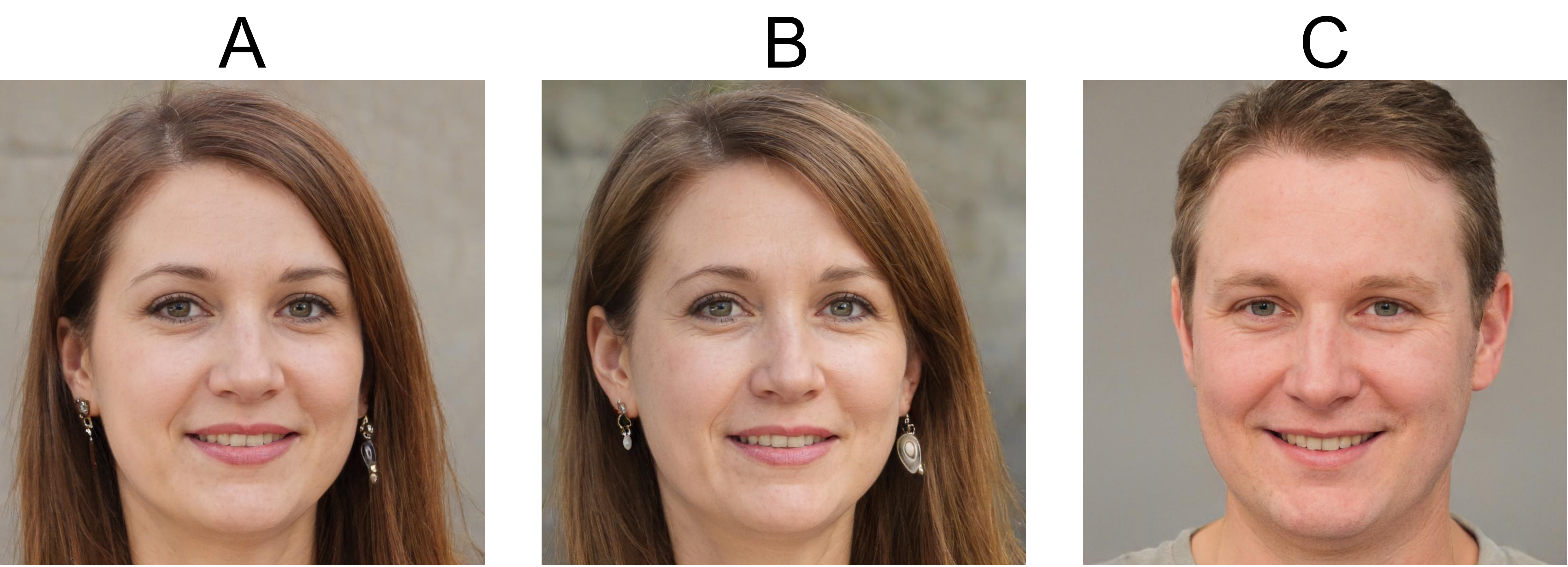}
    \caption{Three example images generated with a state-of-the-art StyleGAN2~\cite{karras2020analyzing} generator. The mean absolute difference between the corresponding latent vectors $w_A$ and $w_B$ is higher compared to the difference between $w_B$ and $w_C$, although images A and B are visually more similar. However, when using face embeddings extracted with ArcFace~\cite{deng2019arcface}, the difference between the face embeddings $e_A$ and $e_B$ is smaller compared to $e_B$ and $e_C$. 
    }
    \label{fig:face_comparison}
\end{figure}

Solely training the reconstruction network to optimise $w_\text{rec}$ does not guarantee that $w_\text{rec} \stackrel{id}{=} w_\text{m}$ because similar latent vectors do not always result in faces that humans perceive as similar. 
For instance, \autoref{fig:face_comparison} shows three faces generated by StyleGAN2. 
Faces A and B appear more visually similar to each other than faces B and C, despite the mean absolute difference between the latent vectors $w_A$ and $w_B$ being larger than between $w_B$ and $w_C$. 
This discrepancy arises because the latent space encodes image features irrelevant to face similarity, such as background, pose, and lighting, independent of what makes faces appear similar to a human observer.

To tackle this issue, we aim to optimise our network based on face embedding vectors instead of the latent vectors directly.
Models like ArcFace~\citep{deng2019arcface} embed faces into a space relevant to identity recognition, as they are specifically trained for this purpose. 
Unlike the latent space $w$, using such face embeddings $e_{A,B,C}$ for the faces in \autoref{fig:face_comparison} results in a mean absolute difference between $e_A$ and $e_B$ that is smaller compared to $e_B$ and $e_C$. 
Consequently, we define the loss function for training the reconstruction network as follows:
\begin{equation}
\mathcal{L} = \|\mathbf{w_\text{rec}} - \mathbf{w_\text{M}}\|^2 - \lambda_\text{e} \frac{E(G(w_\text{rec})) \cdot E(G(w_\text{M}))}{\left|E(G(w_\text{rec}))\right| \left|E(G(w_\text{M}))\right|}
\label{eq_MFRS_loss}
\end{equation}
where $G$ is a pre-trained generator network mapping latent space to image space, $E$ is a pre-trained face embedding network mapping image space to embedding space, and $\lambda_e$ is the embedding similarity loss weighting.
Thus, the reconstruction network aims to predict a latent vector $w_\text{rec}$ similar to the target latent $w_\text{M}$ that also maximises the cosine similarity between the target and predicted face embedding.

During the training of the reconstruction network, we input a variable number of tuples from which it has to infer $w_\text{rec}$.
This allows us to support reconstructing the mental image after any number of iterations without changing the network architecture or loss function as in~\cite{strohm2023usable}, which can negatively impact the reconstruction quality. 
Moreover, this allows us to automatically stop the reconstruction process once no significant changes to the reconstruction can be observed anymore, thus improving usability and reconstruction times:
\begin{equation}
||w_\text{rec}^i-w_\text{rec}^{i+1}||_1 < \alpha,
\label{eq_early_stop}
\end{equation}
where $\alpha$ defines the early termination threshold.
Once the mean absolute difference between two reconstructed latent vectors of consecutive iterations is below this threshold, the iterative process stops, and the latest reconstruction is shown.

\subsection{Computational User Model for Face Similarity Ranking}
\label{sec:user_model}

Training our method end-to-end requires collecting and annotating a costly large-scale dataset. 
For each training sample, a user must memorise a generated face and then rank six auxiliary images over multiple iterations based on their similarity to the memorised face. 
To avoid costly data collection, we introduce a user model that simulates human ranking behaviour, enabling the quick generation of extensive and realistic training data.
The central component of the user model is a face-embedding network that extracts embedding vectors to compute the similarity between two faces. 
We use this network to define the user model described in \autoref{alg:user_model}. 
Given a set of auxiliary faces $\mathcal{F}_\text{aux}^i$, their latent vectors $W_\text{aux}^i$, a target face $f_{m}$, an embedding network $E$, and a noise level $\sigma$, the cosine similarities between the target and auxiliary face embeddings are calculated. 
As can be seen in line six of \autoref{alg:user_model}, we add some noise to the calculated cosine similarities uniformly sampled between $[-\sigma,\sigma]$.
Auxiliary faces that look alike have a comparable cosine similarity with the target face. 
Consequently, adding noise can cause random changes in the ranking of these faces.
The motivation for this arises from the observed noise in user rankings, characterised by significant variability in the rankings assigned by different humans to the same faces.
This observation is discussed in greater detail in \autoref{sec:data-collection}.
Introducing noise into the user model creates a non-deterministic version, which helps to prevent severe overfitting during the optimisation of the reconstruction network.
The auxiliary latent vectors are then sorted according to these noisy similarities, ranking the most similar face first, followed by the others in decreasing order of similarity.

\begin{algorithm}[t]
\caption{User model of face image ranking behaviour: Auxiliary and target faces are projected into an embedding space. The cosine similarity between each auxiliary face embedding and the target face embedding is computed, and random noise is added. These auxiliary face latents are then ordered based on their similarity scores, with the most similar face ranked highest, and the least similar face ranked lowest.}
\label{alg:user_model}
\begin{algorithmic}[1]
\STATE \textbf{Input}: Auxiliary faces for the current iteration $\mathcal{F}_\text{aux}^i$, corresponding latents $W_\text{aux}^i$, \\target face $f_\text{m}$, a face embedding network $E$, and noise level $\sigma$.
\STATE \textbf{Output}: Ranked auxiliary latents $(w_{R1}, w_{R2}, \ldots, w_{Rn}), w_{R} \in W_\text{aux}^i$
\STATE similarities $\gets []$
\FOR{$f$ in $\mathcal{F}_\text{aux}$}
\STATE similarity $\gets \texttt{cosineSimilarity}( E(f), E(f_\text{m}) )$
\STATE noisySim $\gets \texttt{similarity} + \mathcal{U}(-\sigma,\sigma)$
\STATE similarities.append(noisySim)
\ENDFOR
\STATE rankedIndices $\gets \texttt{argSort}(-\text{similarities})$
\STATE rankedLatents $\gets W_\text{aux}^i[\text{rankedIndices}]$
\STATE \textbf{return} rankedLatents
\end{algorithmic}
\end{algorithm}

Existing face embedding models are primarily trained for face identification tasks~\citep{deng2019arcface,wang2018additive,wang2018cosface,parkhi2015deep}. 
Although these models extract meaningful embeddings, they do not explicitly learn to compare and rank faces. 
\citet{sadovnik2018finding} demonstrated that measuring identity does not necessarily equate to measuring similarity, leading to rankings that diverge from human judgement.
To better align the user model with human behaviour, we found that it is beneficial to fine-tune a pre-existing face embedding model on a small face similarity dataset derived from human feedback. 
For fine-tuning, we used a novel dataset we collected (see \autoref{sec:data-collection}) comprising triplets $(f_a,f_p,f_n)$, where $f_a$ is a reference face (anchor), $f_p$ is a face more similar to $f_a$ (positive pair) compared to $f_n$, which is less similar to $f_a$ (negative pair), as determined by human judgement. 
Using this dataset, the embedding network is fine-tuned with a triplet margin loss objective defined as:
\begin{equation}
\mathcal{L}_\text{c} = \max((f_a - f_p)^2 - (f_a - f_n)^2 + m, 0)
\label{eq:triplet}
\end{equation}
where $m$ defines the required margin between the positive and negative pairs to achieve a zero loss. 
This fine-tuning process ensures that the network adjusts the embeddings so that faces perceived as similar by humans are also close in the embedding space.

\subsection{User-based Face Refinement}
\begin{figure}[t]
\centering
\begin{subfigure}[b]{.48\textwidth}
  \centering
  \includegraphics[width=\linewidth]{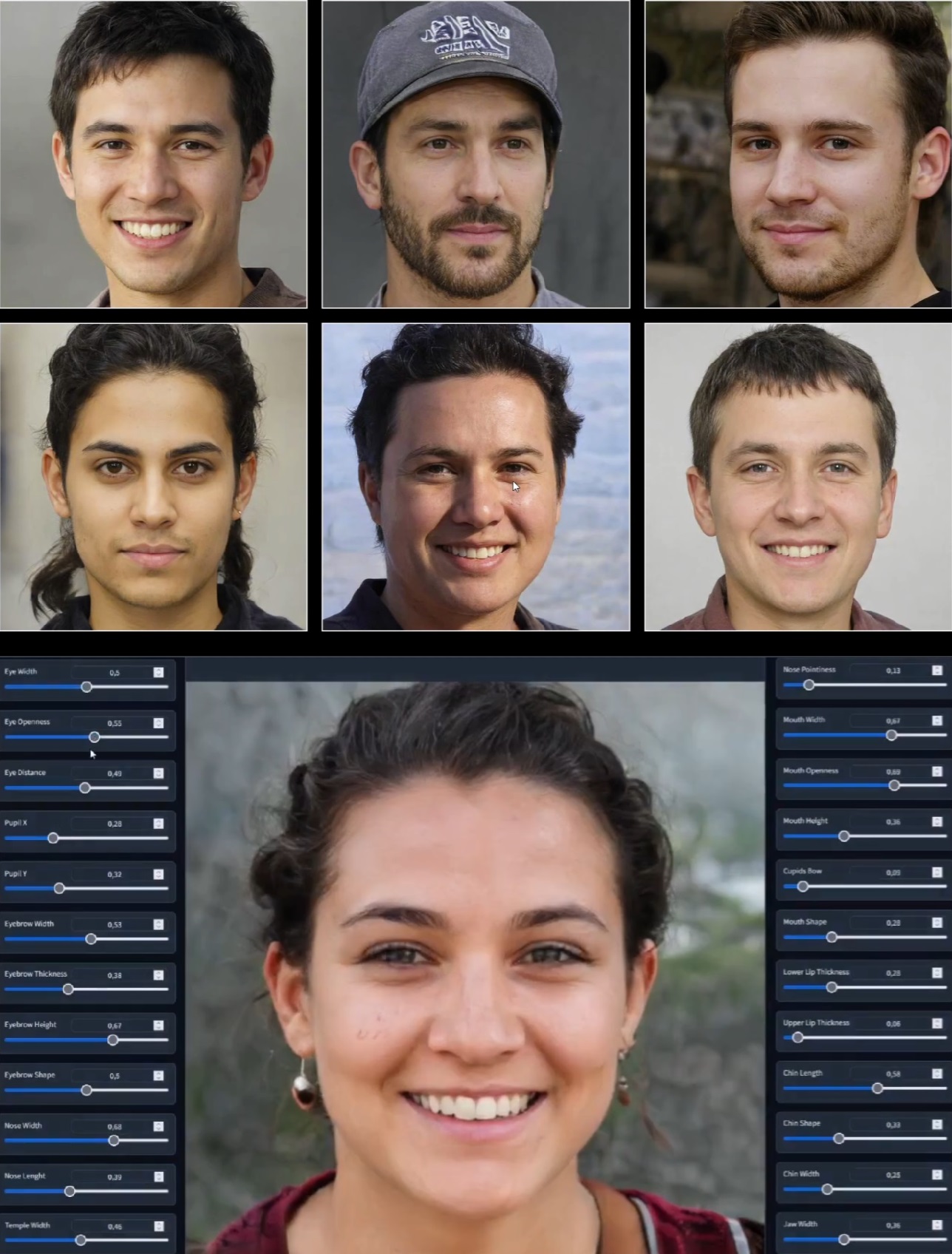}
  \caption{}
  \label{fig:ours_thumbnail}
\end{subfigure}\hfill
\begin{subfigure}[b]{.48\textwidth}
  \centering
  \includegraphics[width=\linewidth]{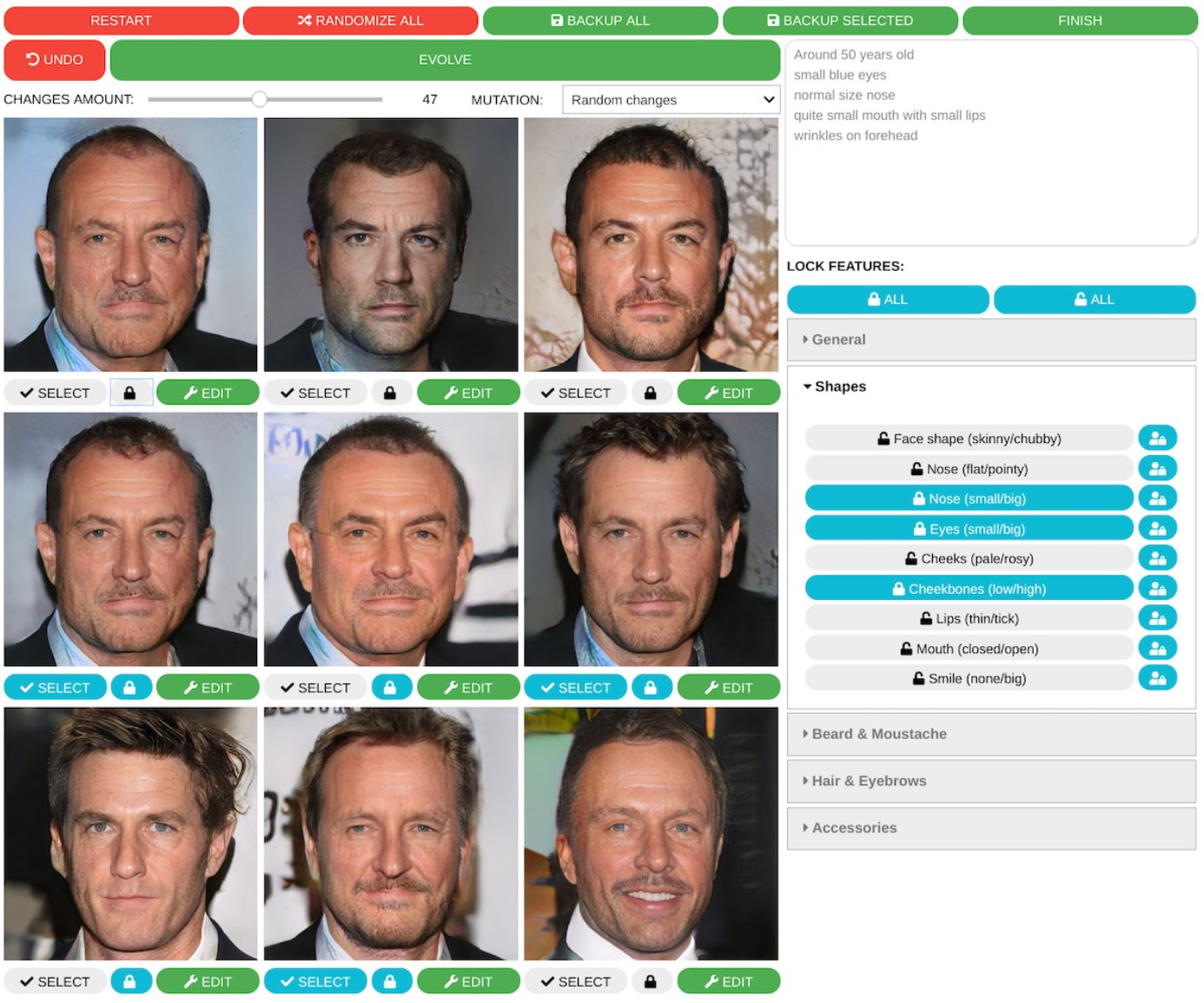}
  \caption{}
  \label{fig:GAN_thumbnail}
\end{subfigure}
\caption{Figure (a) shows the user interface for the first stage of \methodName\ at the top and UP-FacE at the bottom. For our method, users are iteratively presented with six faces that they can rank via drag and drop (top). Afterwards, users are subsequently fine-tuning the reconstruction with UP-FacE (bottom). Figure (b) shows the main interface of CG-GAN. Users are presented with 9 faces that they can manipulate via randomising, mutating or manual editing. The interface contains many buttons to lock specific attribute axes and sub-interfaces for different functionalities like manual editing. }
\label{fig:system-interfaces}
\end{figure}

Upon ranking all tuples of auxiliary images or terminating early based on the criterion defined in \autoref{eq_early_stop}, users are presented with the initial reconstruction generated by \methodName. 
While a holistic approach to face reconstruction is important, prior works like CG-GAN~\cite{zaltron2020cg} have shown that a hybrid approach combining holistic and constructive methods can lead to improved reconstruction results.
Therefore, users can further refine the face manually using UP-FacE~\cite{strohm2024upface}, a tool designed to manipulate face images produced by generative adversarial networks.
Using the same StyleGAN2 model, we can load the initial reconstructed face obtained after the first stage of \methodName\ into UP-FacE. 
This tool offers a user-friendly interface, shown at the bottom of \autoref{fig:ours_thumbnail}, that displays the current image alongside a set of 24 sliders. 
Each slider corresponds to a distinct semantic face feature, such as \textit{nose width} or \textit{chin length}, which users can adjust simply by moving the sliders. 
These 24 semantic face features are defined through 2D facial landmarks, and a pre-trained model has learned to modify the latent vector to reflect the desired changes in the face image.
We opted to integrate UP-FacE into \methodName\ due to its capability to facilitate easy, fine-grained, and precise control over facial features that are vital for face identification, including the eyes, eyebrows, mouth, and nose.
Integrating UP-FacE into \methodName\ allows for a more interactive and precise customisation process, enhancing the overall quality of the final reconstructed face image. 
Users can iteratively adjust the facial features, receiving immediate visual feedback on how each slider manipulation alters the face. 
This interactive refinement ensures that the final output aligns more closely with the user's expectations.
\section{Data Collection}%
\label{sec:data-collection}

To fine-tune the embedding network as described in \autoref{sec:user_model} and to evaluate \methodName\ on real human data, we conducted a data collection user study.

\subsection{Procedure}
The data collection study was conducted online with 408 participants recruited through Amazon's Mechanical Turk (AMT).
Each participant completed 23 trials without time limit, each comprising memorisation and ranking steps. 
During the memorisation phase, we asked participants to look at a random target face generated by StyleGAN2~\citep{karras2020analyzing} until they had memorised it. 
The target face remained consistent throughout the 23 trials, allowing participants to refresh their memory as needed by observing the target face again between trials.
After memorisation, participants were shown each set of auxiliary faces, which matched the sex and age category of the target face, and were instructed to rank the six images based on their perceived similarity to the memorised face. 
Out of the 23 trials, 20 were actual trials, and three were attention checks that we added without telling the participants to ensure data quality.
For these checks, one auxiliary face was replaced with the actual, ground truth target face. 
Proper engagement in the data collection was assumed if participants ranked the target face as the most similar in all three attention checks.

\subsection{Dataset Statistics}
\begin{figure}[t]
\centering
\includegraphics[width=0.6\linewidth]{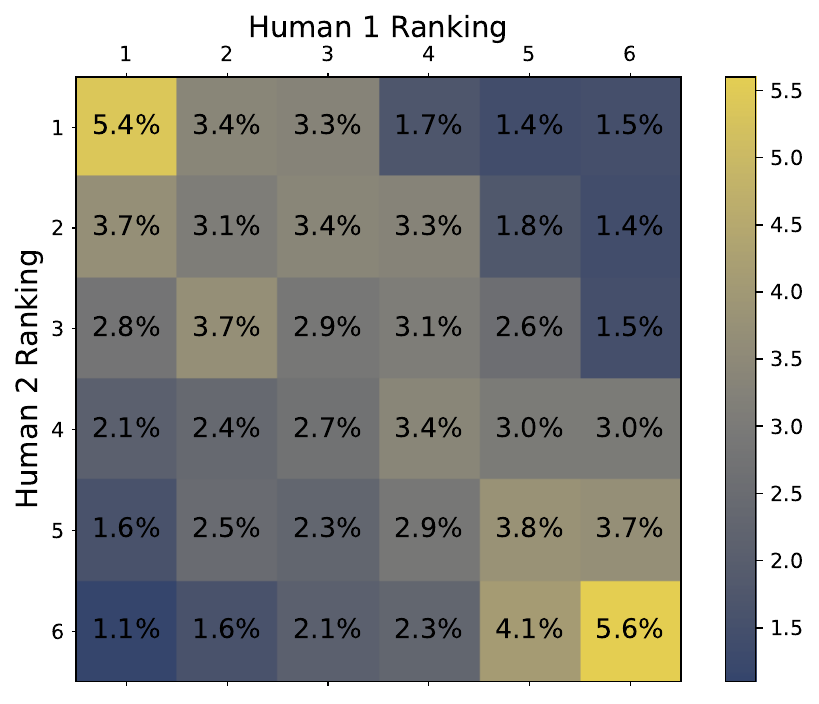}
\caption{Agreement in face rankings among humans is shown in this matrix. Each cell $(i,j)$ represents the probability that two separate human raters will assign ranks $i$ and $j$ to the same face. A positive correlation is noted in human rankings, with an average Kendall's Tau value of 0.267. Additionally, humans tend to show higher consensus on the most and least similar faces, whereas the rankings in the middle range exhibit greater variability.}
\label{fig:human_ranking_agreement}
\end{figure}

We cleaned the dataset by excluding data from participants who failed at least one attention check. 
Out of 408 participants, 76 failed all attention checks, 40 failed two, and 19 failed one, leaving a total of 275 participants in our dataset. 
Rejecting approximately one-third of participants based on standard attention checks is typical for AMT data collections~\cite{saravanos2021hidden}.
Our dataset included 15 target images for which data from two different participants was collected, enabling the calculation of a ranking agreement between participants. 
\autoref{fig:human_ranking_agreement} shows the agreement between two participants for each of the six possible ranks. 
Each cell $(i,j)$ shows the probability that two independent participants assign ranks $i$ and $j$ to the same face. 
The probability of assigning the same ranks is highest, and significant disagreements are rare.
The average Kendall rank correlation coefficient between participants was 0.267 (p<0.05), indicating that humans tend to rank faces similarly, albeit with considerable variability.
This variability in the rankings led to our design choice of a noisy, non-deterministic user model as described in \autoref{sec:user_model}.

We randomly selected 75 out of the remaining 275 participants as a validation set to evaluate the performance of \methodName\ on real human data during training. 
The data from the remaining 200 participants were utilised to generate triplets for fine-tuning the ArcFace~\cite{deng2019arcface} embedding network. 
Out of these, 180 were randomly selected to form the training set, while the data from the remaining 20 participants were used to create the validation set.
For each iteration completed by a participant, we generated $\binom{6}{2}=15$ different $(f_a,f_p,f_n)$ triplets, where $f_a=f_m$ and $(f_p,f_n)$ are all 15 possible pairs of the six ranked auxiliary images. 
The higher-ranked image in a pair was defined as the positive example $f_p$, while the lower-ranked image was the negative example $f_n$.
This resulted in 300 triplets per participant (15 pairs for each of the 20 iterations), yielding a total of 54,000 triplets for training and 6,000 for validation.
\section{Experiments}
\subsection{Implementation Details}

\paragraph{Embedding Network}

For the embedding network in our computational user model and during loss calculation when training the reconstruction network, we used the state-of-the-art face recognition network ArcFace~\cite{deng2019arcface}.
ArcFace was trained on the IBUG-500K dataset that comprises 11.96 million images and 493K identities.
It employs a ResNet50~\cite{he2016deep} neural network architecture as its feature extractor, producing $2048$ $4 \times 4$ feature maps. 
These feature maps are flattened and fed into an output model that includes batch normalisation~\cite{ioffe2015batch}, dropout (with a $40\%$ drop rate)\cite{srivastava2014dropout}, a fully connected layer with 512 neurons, followed by another batch normalisation layer. 
The weights of this model were initialised using the pre-trained ArcFace model. 
We kept the ResNet50 feature extractor frozen while the output model underwent fine-tuning using the collected triplets described in \autoref{sec:data-collection}. 
The network is trained for 100 epochs with a batch size of 32 using the contrastive loss with a margin of 0.1 defined in \autoref{eq:triplet} using an Adam~\cite{kingma2014adam} optimiser with a learning rate of 0.001, $\beta_1 = 0.9$, and $\beta_2 = 0.999$.

\paragraph{Reconstruction Network}
Both transformer encoder modules of the reconstruction network consist of four encoder blocks with hidden dimensions of $512$ and eight attention heads each.
We add sinusoidal positional encodings~\cite{vaswani2017attention} to the input of both transformer modules.
The intermediate Siamese linear and output linear layers consist of 512 neurons without activation function; non-linearities are only present within the transformer modules.
We generated 100K target images for training by randomly sampling latent vectors from a normal distribution and decoding them with StyleGAN2~\cite{karras2020analyzing}. 
Using our sets of auxiliary images and our user model defined in \autoref{alg:user_model} with noise $\sigma = 0.22$, we simulated the human ranking of the auxiliary images for each generated target image.
The reconstruction network can process a variable number of tuples $i\leq20$ containing six ranked latent vectors for each of $i$ iterations. 
We set the maximum number of iterations to 20 and terminated early during test time based on \autoref{eq_early_stop} with $\alpha = 0.1$.
We used the loss function defined in \autoref{eq_MFRS_loss} for training with $\lambda_e=1$.
The model was trained for 100K steps with a batch size of 32 using the Adam optimiser~\cite{kingma2014adam} with a learning rate of 0.0001, $\beta_1 = 0.9$, and $\beta_2 = 0.999$. 
We used the data from the 75 participants left out of the data collection study for validation and selected the model that achieved the lowest validation loss. 

\subsection{Reconstruction Study}
To evaluate our system, we conducted an additional user study with 12 participants (six females) between 24 and 54 years old (Mean=32; SD=12).
Participants were recruited locally among colleagues and friends, and the study design was approved by the university's ethics committee.
After giving informed consent, the study started. 
The study began with explaining the system, allowing participants to familiarise themselves with the process. 
Once participants were comfortable, a target face was presented for memorisation. 
There was no time limit for the memorisation step, but participants could not view the target image again once the experiment had commenced. 
We used the same pool of target faces as in \citet{strohm2023usable}, allowing us to compare our results to theirs.

The reconstructed mental image after the first stage was displayed after completing all 20 ranking iterations or after \methodName\ stopped early.
We then collected data based on the following six evaluation metrics:

\begin{itemize}[leftmargin=*]
    \item \textbf{Mental rating}: Participants rated the similarity between the memorised image and the reconstruction on a seven-point Likert scale, relying solely on their memory of the target image without viewing it during rating.
        
    \item \textbf{Visual rating}: Participants rated the similarity between the target and reconstructed images on a seven-point Likert scale, with both images displayed side by side for comparison.
    
    \item \textbf{Embedding similarity}: We calculated the cosine similarity between the embeddings of the target and reconstructed images to assess the reconstruction quality. As the embedding model is fine-tuned to extract similar embeddings for similar looking faces, this metric allows us to quantitatively compare the reconstruction quality of different methods.

    \item \textbf{Task completion time}: The time participants took to complete the reconstruction. 
    
    \item \textbf{System Usability Scale (SUS)}: Participants completed a SUS questionnaire to assess the system's usability, yielding a score from 0 to 100, with higher scores indicating better usability.
    
    \item \textbf{NASA Task Load Index (NASA-TLX)}: Participants completed the NASA-TLX questionnaire to evaluate perceived workload across six sub-scales: mental demand, physical demand, temporal demand, performance, effort, and frustration. The overall workload score ranges from 0 to 100, with lower scores indicating lower perceived workload.

\end{itemize}

Following the initial reconstruction, participants were asked what changes would make the reconstructed face more similar to their mental image. 
In the second stage, participants used UP-FacE~\cite{strohm2024upface} to refine the initial reconstruction as much as possible.
After this, we again collected data based on the aforementioned six metrics and feedback on possible improvements and tools they would need to enhance the reconstruction further.
The total duration of the study was approximately 30 minutes, depending on how long participants took to complete each step.

\begin{figure*}[t]
    \centering
    \includegraphics[width=\linewidth]{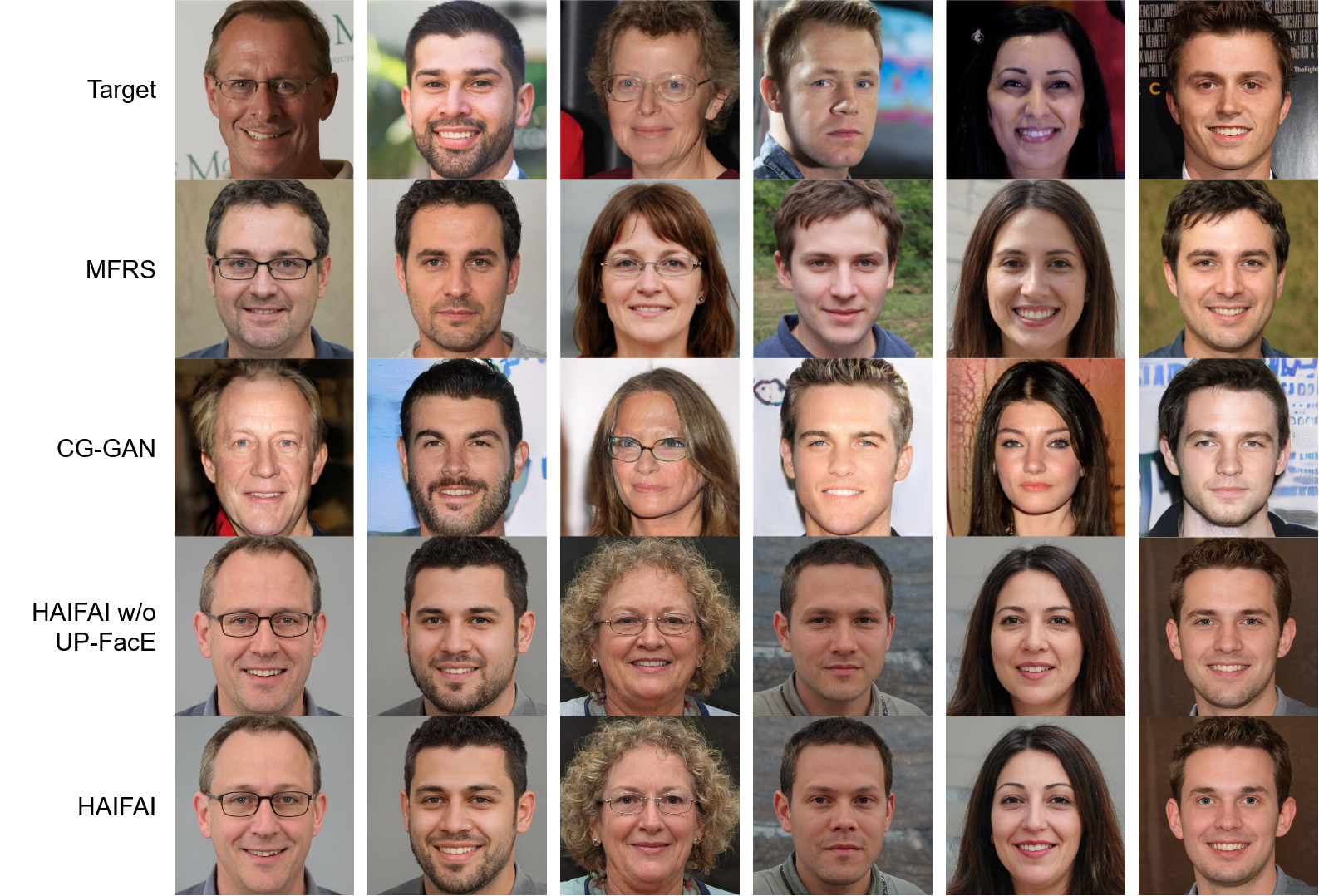}
    \caption{Example reconstructions from our user study compared with the results from \citet{strohm2023usable}. Each column shows the results for one participant. The first row shows the target faces participants had to memorise, and the following two rows show the reconstructions of the baselines. The last two columns show the initial reconstruction from \methodName\ after the first stage as well as the results from \methodName, where faces were further edited with UP-FacE~\cite{strohm2024upface}.}
    \label{fig:user-study-rec}
\end{figure*}

\begin{table}[t]
\begin{center}
\begin{tabularx}{1.0\columnwidth}{l *6{>{\centering\arraybackslash}X}}
\toprule
Method & Mental Rating $\uparrow$ & Visual Rating $\uparrow$ & Embedding Sim. $\uparrow$ & SUS $\uparrow$ & NASA-TLX $\downarrow$ & Time (mins) $\downarrow$\\
\cmidrule(lr){1-1} \cmidrule(lr){2-2} \cmidrule(lr){3-3} \cmidrule(lr){4-4} \cmidrule(lr){5-5} \cmidrule(lr){6-6} \cmidrule(lr){7-7} 
CG-GAN~\cite{zaltron2020cg} & $\mathbf{4.8\phantom{^*} \pm 0.8}$ & $3.9 \pm 0.9$ & $0.36 \pm 0.2$ & $59 \pm 13$ & $43 \pm 11$ & $17.8 \pm 5.6$ \\
MFRS~\cite{strohm2023usable} & $4.0\phantom{^*} \pm 0.8$ & $4.1 \pm 0.9$ & $0.38 \pm 0.1$ & $\underline{85 \pm 13}$ & $\underline{27 \pm 18}$ & $\underline{10.2 \pm 4.1}$  \\
\cmidrule(lr){1-7}
\methodName\ w/o UP-FacE & $4.2^* \pm 0.9$ & $\underline{4.3 \pm 0.8}$ & $\mathbf{0.43^* \pm 0.1}$ & $\mathbf{87 \pm 11}$ & $\mathbf{25 \pm 12}$ & $\mathbf{8.3^* \pm 3.7}$ \\
\methodName & $\underline{4.6\phantom{^*} \pm 0.8}$ & $\mathbf{4.4^* \pm 0.8}$ & $\mathbf{0.43^* \pm 0.1}$ & $77^* \pm 13$ & $31^* \pm 12$ & $11.3 \pm 3.3$ \\
\bottomrule
\end{tabularx}
\end{center}
\caption{Comparison of our proposed method \methodName\ with CG-GAN~\cite{zaltron2020cg} and MFRS~\cite{strohm2023usable} based on our user study results. The best result in each column is highlighted in bold. An asterisk ($^*$) indicates a significant difference to the strongest baseline.}
\label{tbl:results}
\end{table}

%\subsection{Results}
\autoref{fig:user-study-rec} presents example reconstructions from our conducted user study alongside those from two state-of-the-art deep learning based methods: CG-GAN~\cite{zaltron2020cg} and MFRS~\cite{strohm2023usable}. 
CG-GAN is a hybrid reconstruction method that allows users to select and merge faces based on an interactive evolution paradigm, as well as manually edit specific facial attributes.
MFRS is our previous holistic face reconstruction system, which requires users to rank sets of faces similarly to \methodName. 
The top row shows the target image that participants had to memorise. 
The second and third rows display reconstructions produced using the MFRS and CG-GAN methods.
The last two rows present results from our method \methodName\ and an ablated version without UP-FacE~\cite{strohm2024upface}.
Quantitative results from the user study are presented in \autoref{tbl:results}. 
We evaluated the significance of the differences between our methods against the baselines for each metric using either a paired t-test or a Wilcoxon signed-rank test, depending on the normality of the data, as determined by a Shapiro-Wilk test. 
Differences were considered significant if the Bonferroni–Holm corrected p-value was $<0.05$.
An asterisk ($^*$) indicates a significant difference to the strongest baseline, either CG-GAN or MFRS.

Regarding the \textit{mental rating}, participants rated CG-GAN's reconstructions higher than our method's, with scores of 4.8 compared to 4.2 for \methodName\ without UP-FacE and 4.6 with it.
However, our method exceeds the performance of \cite{strohm2023usable}, and there is no statistically significant difference between \methodName\ and CG-GAN.
Regarding the visual rating, \methodName\ outperforms all baselines.
Similar to CG-GAN, we observe a drop from mental to visual rating for \methodName; however, this drop is smaller and not statistically significant, unlike for CG-GAN.
In our newly introduced embedding similarity metric, \methodName\ significantly outperforms the baselines with scores of 0.43, compared to 0.36 for CG-GAN and 0.38 for MFRS.
Further metrics in \autoref{tbl:results} include the average System Usability Scale (SUS) score, NASA Task Load Index (NASA-TLX), and task completion time.
While \methodName\ without UP-FacE outperformed both baselines across these metrics, we can observe a significant performance degradation in these three usability metrics when subsequently using UP-FacE.
The SUS decreases from 87 to 77, NASA-TLX increases from 25 to 31, and the average reconstruction times also significantly increase from 8.3 to 11.3 minutes.

\begin{figure}
    \centering
    \includegraphics[width=\columnwidth]{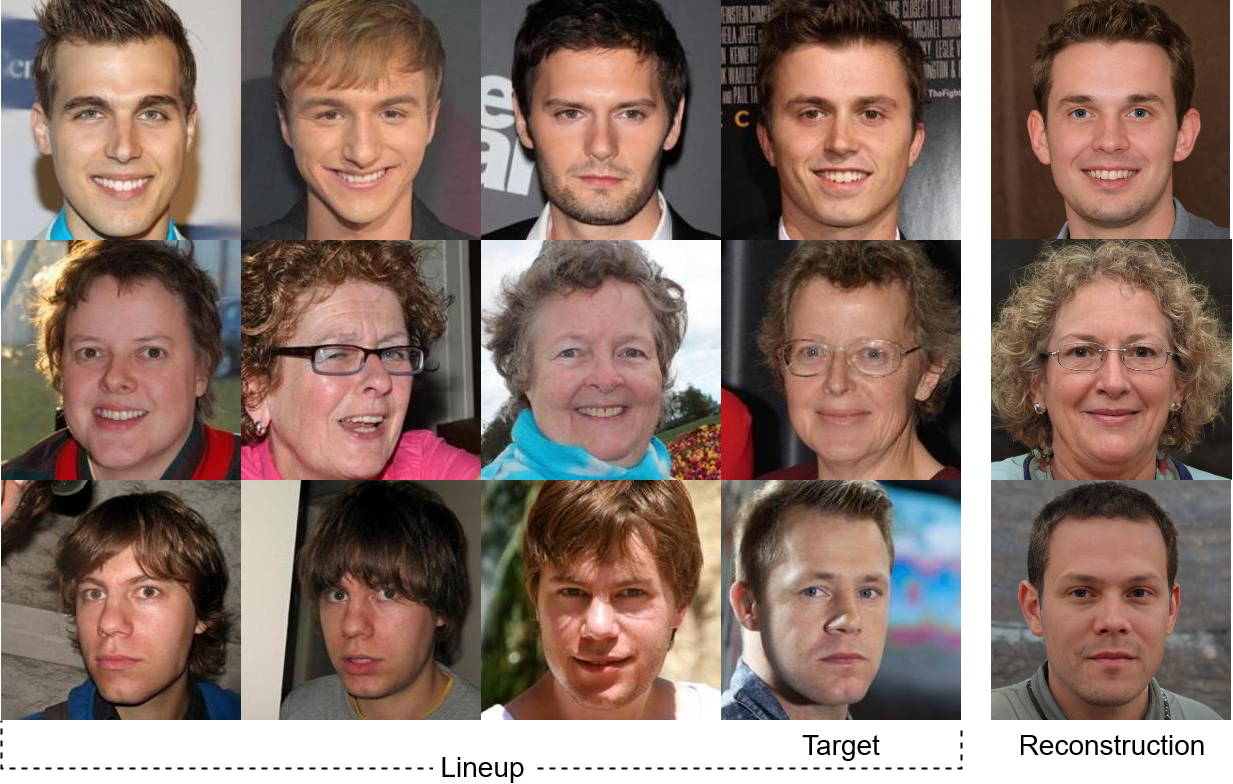}
    \caption{Example lineups used in our lineup study. Participants had to rank the lineup according to the similarity with the reconstruction generated by \methodName.}
    \label{fig:lineups}
\end{figure}

\subsection{Lineup Study}

Beyond the metrics detailed in \autoref{tbl:results}, we performed an additional evaluation to determine the identification rate of our reconstructions through a lineup study, i.e. one of the practical use-cases of our system. 
This is particularly relevant in fields like forensics, where the key objective may not be a flawless reconstruction of the mental image but rather sufficiently good that it leads to correct identification of the individual.
Lineups comprising the true target and similarly appearing faces were created to compute the identification rate. 
Participants were then tasked with ranking these faces based on their resemblance to their reconstruction. 
Following prior work~\cite{zaltron2020cg, strohm2023usable}, the identification rate IR is defined as:

\begin{equation}
\label{eq_recognition_rate}
IR = \frac{\text{\#Rank 1}}{\text{\#Votes}} \times 100.
\end{equation}

The lineup's composition is critical, as an improper selection can skew the results. 
If the faces in the lineup are distinctly different from the target, the identification becomes too easy, artificially boosting the identification rate. 
\citet{zaltron2020cg} addressed this by introducing noise to the latent vectors of generated target faces to create similar-looking faces. 
However, generating such variations is more complex since our targets are real faces.
Instead, we identified the three nearest neighbours within the FFHQ or CelebA-HQ datasets from which the target faces were chosen. 
Using the ArcFace~\cite{deng2019arcface} embedding space, we selected faces with the closest embedding vectors, excluding different images of the same individual. 
This approach yielded 24 lineups, each containing four candidate faces paired with reconstructions from both our methods. 
Example lineups are shown in \autoref{fig:lineups}. 
While this allows us to compare the identification rates of our system with the results from ~\cite{strohm2023usable}, it's important to note that each lineup always includes the target face. 
This condition, which may not hold in real-world situations, could artificially enhance the identification rates for both methods.
We recruited 18 independent raters for an online study. 
Participants were randomly divided into two groups and completed 12 trials in random order. 
Each trial involved ranking a lineup based on similarity to reconstructions generated by \methodName\, either with or without the second stage using UP-FacE, depending on the assigned group. 
This design ensured that each participant evaluated each lineup only once, mitigating potential biases and enabling cross-group comparison of all reconstructions.
The study results showed an identification rate IR of $60.6\%$ for \methodName\ and $59.3\%$ without the second stage. 
Both of these results mark a statistically significant improvement to the previous leading performance of $56.1\%$ by CG-GAN, as determined by a Wilcoxon signed-rank test with $p<0.05$.
Moreover, participants could rank the target faces within the top three based on our systems reconstructions in $98.5\%$ of the cases, improving over the 95.0\% previously achieved by CG-GAN.

\subsection{Ablation Experiments}
\begin{figure}[t]
    \centering
    \includegraphics[width=\linewidth]{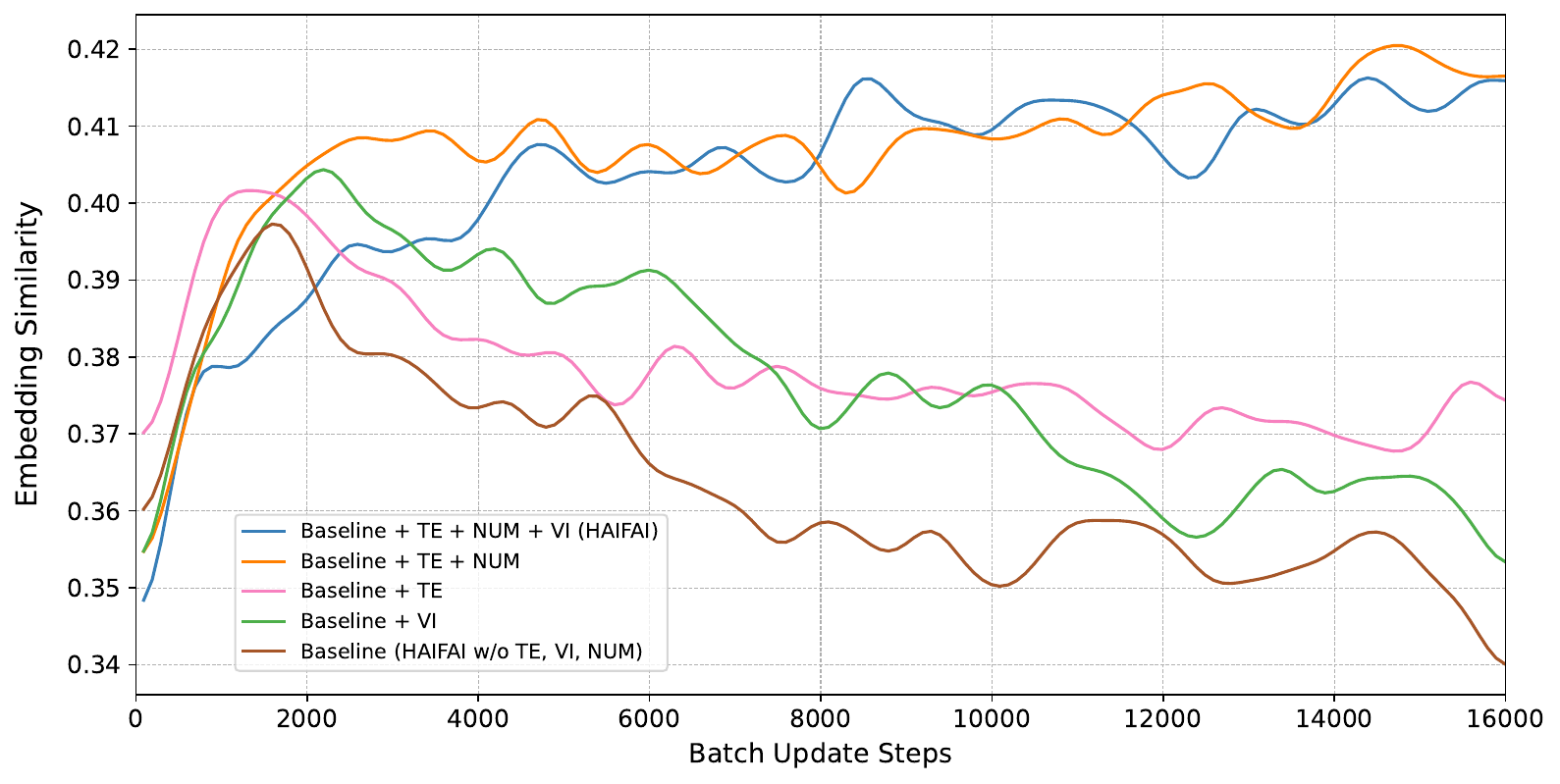}
    \caption{The figure presents a plot illustrating the embedding similarity (where higher values indicate better performance) as a function of the number of batch update steps during training for various ablated models. The brown baseline curve represents the performance of the first stage of \methodName, but with a deterministic user model, without fine-tuned embeddings, and without variable iterations. It is evident that incorporating variable iterations (VI), fine-tuned embeddings (TE), and a noisy user model (NUM) each contributes to improving the overall model performance.}
    \label{fig:ablation}
\end{figure}

We conducted a series of ablation experiments to evaluate the effectiveness of specific model design decisions. 
\autoref{fig:ablation} shows Gaussian-smoothed curves for five different models, with the number of batch update steps during training on the x-axis and test-set embedding similarity on the y-axis (higher is better).
The baseline model, represented by the brown curve, corresponds to the first stage of \methodName\ without incorporating fine-tuned embeddings (TE), variable iterations (VI), or the noisy user model (NUM).
This model achieves a peak test-set embedding similarity of 0.397 but begins to severely overfit thereafter, as indicated by the decline in test-set embedding similarity with additional updates.
The pink line illustrates the performance of the baseline model when using the embedding network fine-tuned on our collected data, as described in \autoref{sec:data-collection}. 
We observe a slight improvement in peak embedding similarity to 0.402 and reduced overfitting when utilising fine-tuned embeddings.
The green line represents the results for the baseline model when training with a variable input sequence length instead of a fixed 20 iterations as in \cite{strohm2023usable}. 
This approach not only permits early termination of the reconstruction process, as described in \autoref{sec:method}, but also appears to regularise the network, resulting in an improved embedding similarity of 0.405 and diminished overfitting.
The most significant performance enhancement is achieved by introducing noise into the user model, as described in \autoref{alg:user_model}. 
This is evidenced by the orange and blue lines, which no longer overfit the training data and reach a peak embedding similarity of 0.421.
The blue line, representing the first stage of \methodName\ used in our evaluation studies, demonstrates that training with a variable number of iterations as input does not negatively impact reconstruction quality. However, it requires a longer training duration to reach a comparable performance.
Unlike \cite{strohm2023usable} this allows us to stop the reconstruction at anytime without reconstruction quality degradation when using all iterations.

\begin{figure}[t]
    \centering
    \includegraphics[width=\linewidth]{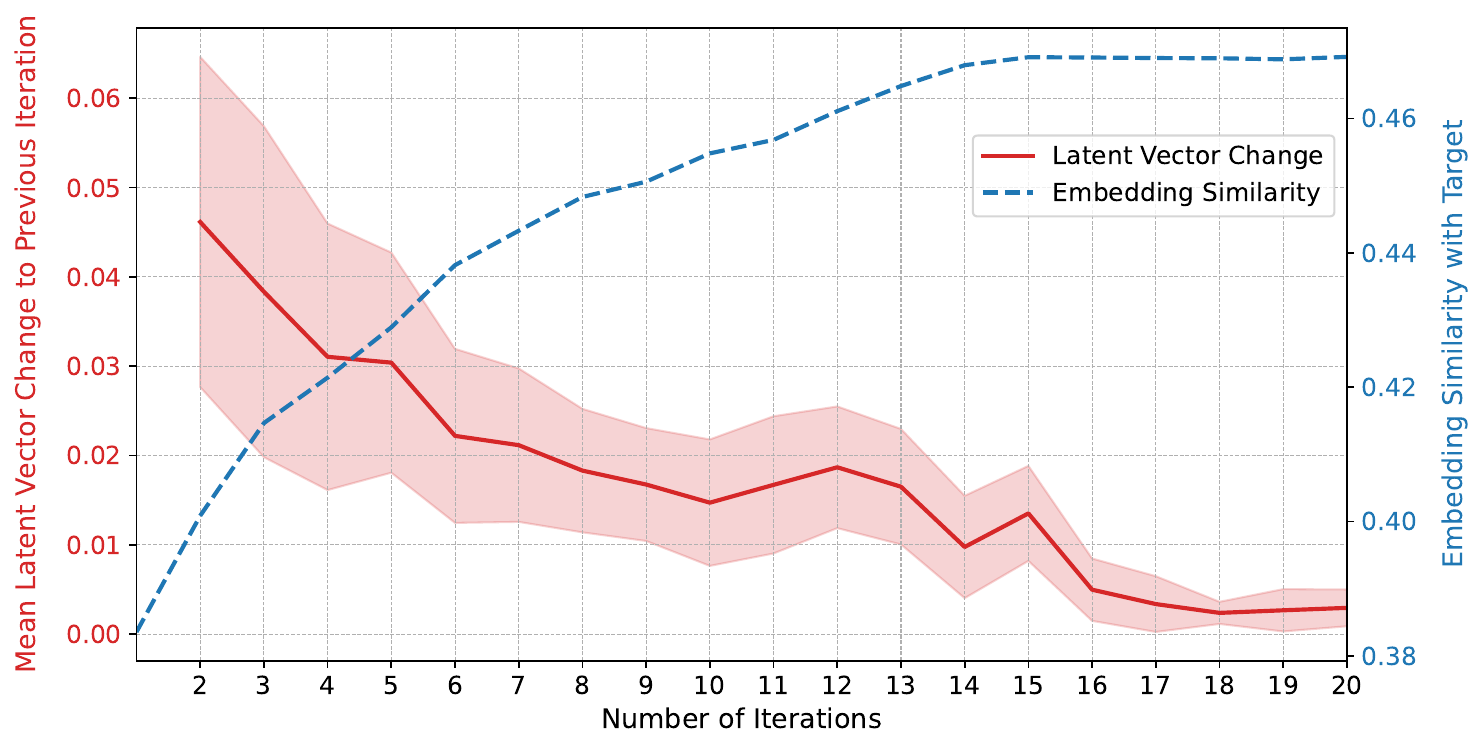}
    \caption{This figure shows the change in latent vector (red line) as well as the embedding similarity (blue line) over the number of iterations used for reconstructing the image. We observe that the model converges after 15 iterations on average.}
    \label{fig:num_iters}
\end{figure}

\paragraph{Number of Iterations}
The maximum number of iterations for \methodName\ is set to 20, as additional iterations do not further enhance the reconstructions. 
\autoref{fig:num_iters} illustrates the number of iterations on the x-axis, with the mean absolute change of the latent vector compared to the previous iteration on the left y-axis, and the validation set embedding similarity on the right y-axis.
As anticipated, the embedding similarity increases monotonically with the number of iterations until it begins to converge around 15 iterations. 
This trend is consistent with the observed change in the latent vector, which approaches zero after 15 iterations. 
These observations suggest that, on average, the reconstruction process can be effectively terminated after 15 iterations without compromising reconstruction quality.
Based on the observations made in \autoref{fig:num_iters}, we set the early termination threshold $\alpha$ for \autoref{eq_early_stop} to 0.1.
The reduction in required iterations through automatic early termination is evident in the decreased reconstruction time presented in \autoref{tbl:results}, compared to \cite{strohm2023usable}, which utilised a fixed 20 iterations.
During our user study, our system terminated after as few as 10 iterations and never exceeded 19 iterations.
The stopping criterion might be too aggressive, as it immediately halts after no change is observed for a single iteration.
\section{Discussion}
\subsection{Comparison with Baselines}
\label{sec:baseline_comp}

Our user study results (\autoref{tbl:results}) revealed substantial improvements in all six metrics for \methodName\ without the second stage compared to the previous state-of-the-art from \citet{strohm2023usable}.
While \methodName\ with the second stage achieves the best reconstruction quality based on the user ratings, it comes at the cost of slightly higher workload, reconstruction times, and lower usability.
To evaluate reconstruction quality, we used two user-based metrics: \textit{mental rating} and \textit{visual rating}. 
The \textit{mental rating} gauges users' perceived similarity to the target image based on their memory at the end of the experiment, whereas the \textit{visual rating} involves a side-by-side comparison of the target image and the reconstruction, allowing users to more accurately assess reconstruction quality.
Although CG-GAN achieved a higher mental rating (4.8) than our method (4.2/4.6), \methodName\ surpassed both CG-GAN and \citet{strohm2023usable} in the visual rating. 
Notably, CG-GAN’s score dropped more substantially from mental (4.8) to visual (3.9) rating, while \methodName’s drop was smaller (4.6 to 4.4) and not statistically significant. 
This discrepancy between mental and visual ratings may stem from participants' mental image shifting toward the reconstructed face during editing, a phenomenon we discuss in detail in \autoref{sec:discussion_humanfactors}.
In many applications, particularly in forensics, the primary objective is not necessarily to maximize perceived similarity but rather to enhance the probability of correctly identifying the target from the reconstruction. 
Our additional user study found that \methodName\ achieves an identification rate of $60.6\%$, which is notably higher than CG-GAN’s $56.1\%$. 
Given the large sizes of the CelebA-HQ~\cite{karras2017progressive} and FFHQ~\cite{karras2019style} datasets (30K and 70K images, respectively), and our approach of selecting nearest neighbours for challenging lineups, this level of performance is encouraging. 
Furthermore, in $98.5\%$ of cases, the target face was not ranked last, showing our reconstructions were closer to the true target than at least one of the three nearest neighbours. 
Interestingly, despite CG-GAN’s higher mental rating, our method’s higher identification rate suggests the mental rating metric may be skewed by factors unrelated to true similarity.
Beyond reconstruction quality, our methods significantly outperform CG-GAN on three key usability metrics: users gave higher SUS scores, lower NASA-TLX ratings, and reported faster task completion times. 
We attribute these improvements to a more streamlined, ranking-based approach that shifts the complexity of high-dimensional face search from the user to the system.

\subsection{Human Factors and Interaction Design} 
\label{sec:discussion_humanfactors}

Although the core contribution of \methodName\ is computational, our user studies and participant feedback highlight important human factors and interaction design insights for future mental image reconstruction methods. 

\paragraph{Minimizing Cognitive Load.}
Compared to more feature-intensive systems like CG-GAN \cite{zaltron2020cg}, our ranking-based approach simplifies interaction and reduces cognitive load. 
In CG-GAN, participants must choose a suitable base face, explore multiple parallel feature spaces, and refine attributes individually. 
This complexity was reflected in CG-GAN’s significantly higher NASA-TLX score of 43.4 compared to \methodName’s 27.2 (\autoref{tbl:results}). 
By contrast, \methodName\ decomposes the reconstruction task into straightforward ranking steps, after which the system takes on the challenge of high-dimensional exploration. 
This approach aligns with human cognitive patterns, where broad judgments (e.g., “this face is closer to what I remember”) are more easily handled than incremental tuning of many separate attributes.

\paragraph{Structured Task Allocation.} 
A key insight of our work is the clear benefit of a structured, two-stage interaction approach. 
In the first stage, users only rank sets of candidate faces on overall similarity, offloading the heavy lifting to the AI system, which integrates this feedback holistically. 
Our participants noted that these small, discrete tasks were less overwhelming and helped maintain a stable mental image. 
This design contrasts with purely manual editing, where the user must recall many details simultaneously (e.g., nose shape, eye spacing, brow curvature), often leading to frustration. 
From an interaction-design perspective, our results suggest that incremental user input, such as short ranking tasks, can outperform extensive manual editing for tasks requiring high-quality reconstructions with minimal user burden.
Moreover, our SUS results indicate that users appreciate a balanced \emph{division of labour} between themselves and the AI. 
With \methodName, the AI quickly identifies a roughly correct latent space region. 
The user only steps in at a fine-grained level when they feel confident in adjusting specific details. 
Interviews and open-text feedback further underscored that participants valued having a manageable number of actions per stage. 
They felt a sense of agency in the second stage, which is still essential for personal satisfaction, yet welcomed the system's autonomy to handle global exploration. 
In this sense, \methodName\ underscores a broader principle: tasks that AI can do quickly and more systematically, like searching for a rough match in a high-dimensional space, are best left to the AI, whereas humans excel at noticing subtle details and making targeted refinements once a decent approximation is available.

\paragraph{Manual Editing.} 
Although the ranking-based reconstruction alone already achieved strong performance, many participants felt it was \emph{``incomplete''} when it came to final subtle adjustments, particularly for facial features that are highly important and define user identity (eyes, nose, mouth). 
By integrating UP-FacE~\cite{strohm2024upface} as a second stage, \methodName\ becomes a hybrid system that fuses holistic and constructive methods. 
This inclusion led to improvements in the user-rated similarity, embedding similarity, and line-up identification rates, albeit with some trade-offs:
In our study, time to completion increased from 8.3 to 11.3 minutes, and usability ratings dropped from 87 to 77 on the SUS (\autoref{tbl:results}). 
These results indicate a trade-off: the potential for higher-quality outcomes versus additional user burden and time. 
In domains like forensics, where accuracy might outweigh ease and speed, participants generally found the second stage worthwhile. 
When polled on what they wished to modify after the system’s initial reconstruction, most participants focused on the most important features: eyes, nose, mouth, and general head shape. 
Secondary interests included hairstyle and colour, beard, age, skin tone, and head orientation. 
Although these secondary features can be changed in real life (e.g., hairstyle), many users still perceived them as relevant for perceived facial similarity.
Some participants suggested that future systems might benefit from lightweight text-based editing (e.g., ``make the hair shorter'' or ``add a beard'') to handle these changeable properties more intuitively. 
However, participants also expressed concerns about interface complexity, noting that while having more means to modify the face is beneficial, these options should be hidden or optional to avoid causing confusion.
Our study results also suggest that some participants would value built-in ``smart suggestions'' based on their ranking patterns, e.g., automatic age adjustments or demographic-based style changes if consistently selected. 
Such adaptivity and deeper mutual understanding between the user and AI could further streamline the process. 

Interestingly, the extent of \textit{active} manual editing appears linked to a phenomenon referred to as ``mental shift''.
Previous work by \citet{strohm2023usable} identified a significant drop from mental to visual ratings for CG-GAN~\cite{zaltron2020cg}, where manual editing is central.
They attributed it partly to participants unintentionally internalising the edited image as their mental reference. 
Using their own method, however, this effect was not present, likely because their method does not require active editing. 
This suggests that methods requiring fewer manual interactions help maintain a more stable memory of the target.
Likewise, our current results show that after the first stage, which involves no active face editing, there is even a slight (albeit not statistically significant) increase from mental to visual rating. 
However, after the second stage, where participants can manually edit face features, we again observe a drop in ratings. Still, it is notably smaller than in CG-GAN, likely because the necessary adjustments in \methodName\ are far less extensive. 
These findings imply that reducing manual editing demands is beneficial for tasks relying on an intact mental image. 
A higher degree of active editing can inadvertently shift a person’s memory, highlighting a human-factors challenge that goes beyond purely computational concerns. 
In high-stakes use cases like forensic reconstructions, where accurate memory is paramount, systems should aim to minimise the user’s manual intervention or at least offer a structured, short-burst approach to preserve the fidelity of the mental image.

\begin{figure*}[t]
\centering
\begin{subfigure}{.5\textwidth}
  \centering
  \includegraphics[width=\linewidth]{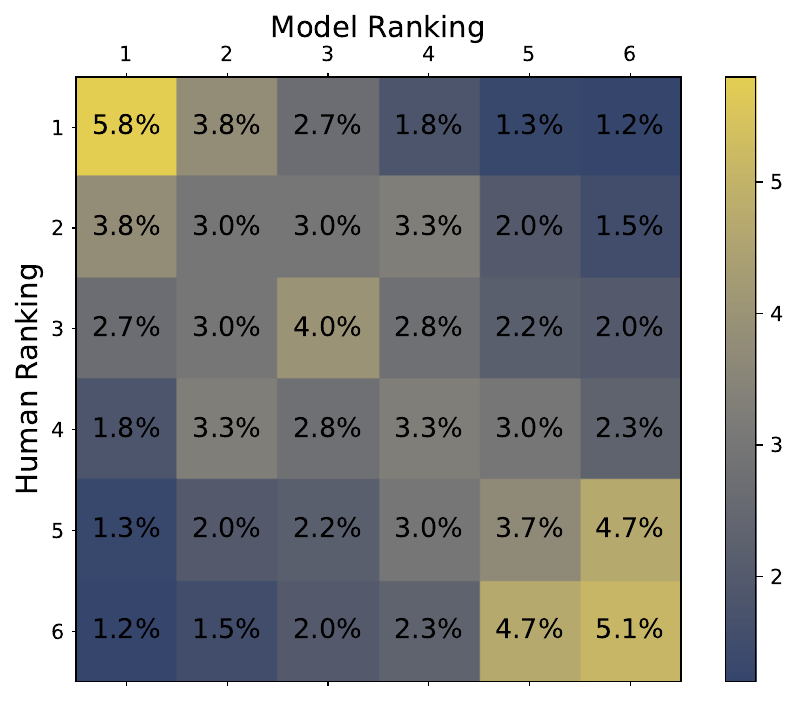}
  \caption{}
  \label{fig:original-model_ranking_agreement}
\end{subfigure}%
\begin{subfigure}{.5\textwidth}
  \centering
  \includegraphics[width=\linewidth]{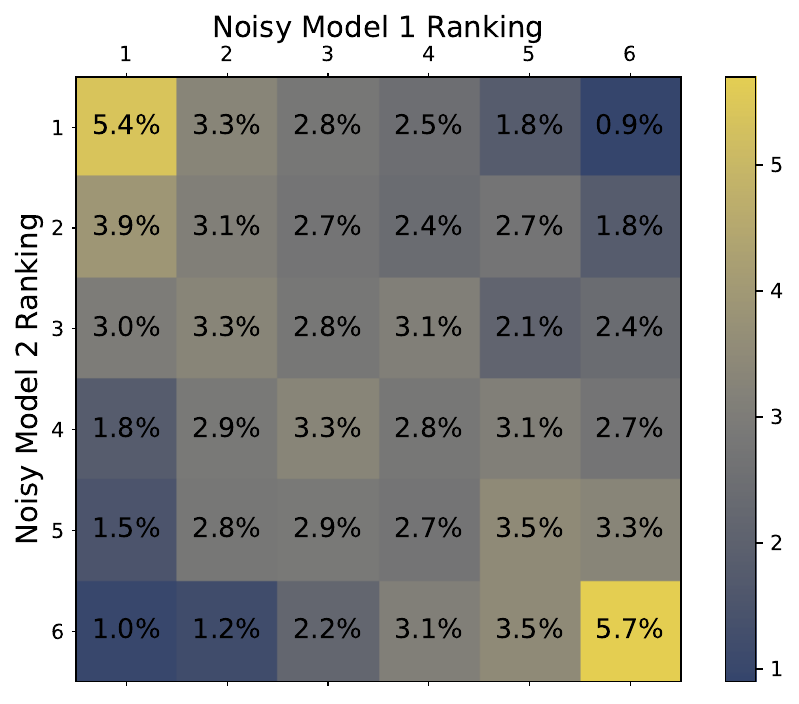}
  \caption{}
  \label{fig:noisy-model_ranking_agreement}
\end{subfigure}
\caption{Face ranking agreement between humans and our user model (a) and agreement between the noisy user model itself (b). For the latter, we generated two rankings by sampling from the noisy user model and compared these rankings to assess the internal agreement. Each cell $(i,j)$ shows the probability that human raters/user model assign rank $i$ and the user model rank $j$ to the same face.}
\label{fig:model-human-ranking}
\end{figure*}

\subsection{Noisy User Model}
Since we are training \methodName\ with simulated human data, achieving high similarity between real human rankings and simulated rankings is essential. 
\autoref{fig:original-model_ranking_agreement} displays a $6 \times 6$ matrix indicating the agreement between our user model and human rankings: each cell $c_{i,j}$ shows the probability that a human assigns rank $i$ to an image. In contrast, the model assigns rank $j$.
Compared to the human ranking agreement in \autoref{fig:human_ranking_agreement}, our computational user model in \autoref{fig:original-model_ranking_agreement} demonstrates rankings that closely align with human judgments. 
While the human-human Kendall rank correlation coefficient is $0.267$, the model-human coefficient is $0.284$ (p < 0.05), indicating similar average ranking behaviour.
We also investigated why injecting noise improves model performance (\autoref{fig:ablation}). 
A deterministic user model yields a model-model rank correlation of $1.0$ by definition and leads to severe overfitting: the network essentially learns to perfectly match an artificially clean distribution, which poorly generalises to real-world data. 
By injecting noise into the embedding similarities as described in \autoref{alg:user_model}, we obtain noisy rankings more akin to human variation. 
We determined the noise level $\sigma = 0.22$ via grid search, minimizing the Wasserstein distance between the human-human and model-model ranking distributions. 
This results in the model-model ranking distribution shown in \autoref{fig:noisy-model_ranking_agreement}, with a Kendall rank correlation of $0.258$, which is close to the human-human correlation, thereby making the simulated rankings more realistic.

An important takeaway from our work is that an improved user model improves reconstruction performance. 
While our evaluation suggests that the current user model already approximates average human ranking behaviour well, as evidenced by the similar human-human and model-human correlation coefficients, there remains potential for further refinement. 
Human rankings exhibit a degree of noisiness on average, but it might be possible to reduce this variance by better modelling individual user preferences. 
For instance, instead of using a static user model that represents the average human, a dynamic user model could adapt to the specific ranking behaviour of individual users. 
This personalised approach may help reduce noise in the simulated rankings, offering the reconstruction network richer and more meaningful signals during training. 
As a result, it could further enhance reconstruction performance while ensuring robust generalisation to unseen users.

\subsection{Limitations and Future Work} \label{sec:limitations}
A practical limitation of \methodName\ arises not from our method itself but from the biases present in the state-of-the-art generative models on which it builds. 
Specifically, the StyleGAN backbone used in our approach is trained on well-known facial image datasets (e.g., FFHQ~\cite{karras2019style}, CelebA-HQ~\cite{karras2017progressive}) that do not fully capture the global diversity of facial features. 
Consequently, these datasets introduce biases in the generated faces, particularly if certain racial or ethnic attributes are under-represented.
Due to these biases, our current work has focused primarily on Caucasian faces. 
However, with a more balanced generative model, our method could easily be extended to include an initial option for selecting a preferred ethnicity, enabling the system to generate and display matching auxiliary outputs accordingly. 
Moreover, while \methodName\ offers strong baseline performance, certain applications may require even more sophisticated editing of external and stylised attributes (e.g., hairstyle, lighting, accessories). 
Although UP-FacE already allows refined control of facial features, further integration of attribute-based or language-based editing tools could improve realism at the cost of more interface complexity and user effort.
Finally, \methodName\ was tested under controlled lab conditions with \emph{new} faces that participants had only briefly studied. 
In real-world settings, e.g., generating a facial composite of someone a witness saw weeks before, the user’s memory could be incomplete or less accurate. 
Investigating how different levels of familiarity and recall difficulty impact ranking consistency and final reconstructions is an open area for future work.
\section{Conclusion}

In this work we introduced \methodName\ -- a novel system where human and AI interact to reconstruct the mental image of the user. 
Unlike previous methods that required users to reconstruct mental images using cumbersome and time-intensive tools, our approach only requires users to iteratively rank images of faces based on their similarity to their mental image. 
Our system integrates image features across all iterations to visually decode the mental image using a state-of-the-art generative model.
In a further step users can optionally further improve the reconstruction manually with an easy-to-use slider interface.
We validated our system through extensive quantitative evaluations involving two user studies with a total of 30 participants. 
The results of these studies showed our system's superior performance in terms of reconstruction quality and time, identification rate, usability, and cognitive load.
These findings underscore the potential of human-AI interaction in enhancing cognitive tasks, paving the way for future advancements in personalised AI applications. Further research could explore extending this approach to other domains where mental imagery and subjective experience play a crucial role.

\bibliographystyle{ACM-Reference-Format}
\bibliography{sample-base}


\begin{thebibliography}{101}

%%% the Web address.

\ifx \showCODEN    \undefined \def \showCODEN     #1{\unskip}     \fi
\ifx \showDOI      \undefined \def \showDOI       #1{#1}\fi
\ifx \showISBNx    \undefined \def \showISBNx     #1{\unskip}     \fi
\ifx \showISBNxiii \undefined \def \showISBNxiii  #1{\unskip}     \fi
\ifx \showISSN     \undefined \def \showISSN      #1{\unskip}     \fi
\ifx \showLCCN     \undefined \def \showLCCN      #1{\unskip}     \fi
\ifx \shownote     \undefined \def \shownote      #1{#1}          \fi
\ifx \showarticletitle \undefined \def \showarticletitle #1{#1}   \fi
\ifx \showURL      \undefined \def \showURL       {\relax}        \fi
% invisible to TeX
\providecommand\bibfield[2]{#2}
\providecommand\bibinfo[2]{#2}
\providecommand\natexlab[1]{#1}
\providecommand\showeprint[2][]{arXiv:#2}

\bibitem[Abdal et~al\mbox{.}(2021)]%
        {abdal2021styleflow}
\bibfield{author}{\bibinfo{person}{Rameen Abdal}, \bibinfo{person}{Peihao Zhu}, \bibinfo{person}{Niloy~J Mitra}, {and} \bibinfo{person}{Peter Wonka}.} \bibinfo{year}{2021}\natexlab{}.
\newblock \showarticletitle{Styleflow: Attribute-conditioned exploration of stylegan-generated images using conditional continuous normalizing flows}.
\newblock \bibinfo{journal}{\emph{ACM Transactions on Graphics (ToG)}} \bibinfo{volume}{40}, \bibinfo{number}{3} (\bibinfo{year}{2021}), \bibinfo{pages}{1--21}.
\newblock


\bibitem[Arnheim(1969)]%
        {arnheim2023visual}
\bibfield{author}{\bibinfo{person}{Rudolf Arnheim}.} \bibinfo{year}{1969}\natexlab{}.
\newblock \bibinfo{booktitle}{\emph{Visual thinking}}.
\newblock \bibinfo{publisher}{Univ of California Press}.
\newblock


\bibitem[Beliy et~al\mbox{.}(2019)]%
        {beliy2019voxels}
\bibfield{author}{\bibinfo{person}{Roman Beliy}, \bibinfo{person}{Guy Gaziv}, \bibinfo{person}{Assaf Hoogi}, \bibinfo{person}{Francesca Strappini}, \bibinfo{person}{Tal Golan}, {and} \bibinfo{person}{Michal Irani}.} \bibinfo{year}{2019}\natexlab{}.
\newblock \showarticletitle{From voxels to pixels and back: Self-supervision in natural-image reconstruction from fMRI}. In \bibinfo{booktitle}{\emph{Advances in Neural Information Processing Systems}}. \bibinfo{pages}{6517--6527}.
\newblock


\bibitem[Bontrager et~al\mbox{.}(2018)]%
        {bontrager2018deep}
\bibfield{author}{\bibinfo{person}{Philip Bontrager}, \bibinfo{person}{Wending Lin}, \bibinfo{person}{Julian Togelius}, {and} \bibinfo{person}{Sebastian Risi}.} \bibinfo{year}{2018}\natexlab{}.
\newblock \showarticletitle{Deep interactive evolution}. In \bibinfo{booktitle}{\emph{International Conference on Computational Intelligence in Music, Sound, Art and Design}}. \bibinfo{pages}{267--282}.
\newblock


\bibitem[Bruera and Poesio(2022)]%
        {bruera2022exploring}
\bibfield{author}{\bibinfo{person}{Andrea Bruera} {and} \bibinfo{person}{Massimo Poesio}.} \bibinfo{year}{2022}\natexlab{}.
\newblock \showarticletitle{Exploring the representations of individual entities in the brain combining EEG and distributional semantics}.
\newblock \bibinfo{journal}{\emph{Frontiers in artificial intelligence}} (\bibinfo{year}{2022}), \bibinfo{pages}{25}.
\newblock


\bibitem[Bulling and Roggen(2011)]%
        {bulling11_ubicomp}
\bibfield{author}{\bibinfo{person}{Andreas Bulling} {and} \bibinfo{person}{Daniel Roggen}.} \bibinfo{year}{2011}\natexlab{}.
\newblock \showarticletitle{Recognition of Visual Memory Recall Processes Using Eye Movement Analysis}. In \bibinfo{booktitle}{\emph{Proc. ACM International Joint Conference on Pervasive and Ubiquitous Computing (UbiComp)}}. \bibinfo{pages}{455--464}.
\newblock
\urldef\tempurl%
\url{https://doi.org/10.1145/2030112.2030172}
\showDOI{\tempurl}


\bibitem[Chen et~al\mbox{.}(2021)]%
        {chen2021deepfaceediting}
\bibfield{author}{\bibinfo{person}{Shu-Yu Chen}, \bibinfo{person}{Feng-Lin Liu}, \bibinfo{person}{Yu-Kun Lai}, \bibinfo{person}{Paul~L Rosin}, \bibinfo{person}{Chunpeng Li}, \bibinfo{person}{Hongbo Fu}, {and} \bibinfo{person}{Lin Gao}.} \bibinfo{year}{2021}\natexlab{}.
\newblock \showarticletitle{DeepFaceEditing: deep face generation and editing with disentangled geometry and appearance control}.
\newblock \bibinfo{journal}{\emph{ACM Transactions on Graphics (TOG)}} \bibinfo{volume}{40}, \bibinfo{number}{4} (\bibinfo{year}{2021}), \bibinfo{pages}{1--15}.
\newblock


\bibitem[Chen et~al\mbox{.}(2020)]%
        {chen2020deepfacedrawing}
\bibfield{author}{\bibinfo{person}{Shu-Yu Chen}, \bibinfo{person}{Wanchao Su}, \bibinfo{person}{Lin Gao}, \bibinfo{person}{Shihong Xia}, {and} \bibinfo{person}{Hongbo Fu}.} \bibinfo{year}{2020}\natexlab{}.
\newblock \showarticletitle{DeepFaceDrawing: Deep generation of face images from sketches}.
\newblock \bibinfo{journal}{\emph{ACM Transactions on Graphics (TOG)}} \bibinfo{volume}{39}, \bibinfo{number}{4} (\bibinfo{year}{2020}), \bibinfo{pages}{72--1}.
\newblock


\bibitem[Chiu et~al\mbox{.}(2020)]%
        {chiu2020human}
\bibfield{author}{\bibinfo{person}{Chia-Hsing Chiu}, \bibinfo{person}{Yuki Koyama}, \bibinfo{person}{Yu-Chi Lai}, \bibinfo{person}{Takeo Igarashi}, {and} \bibinfo{person}{Yonghao Yue}.} \bibinfo{year}{2020}\natexlab{}.
\newblock \showarticletitle{Human-in-the-loop differential subspace search in high-dimensional latent space}.
\newblock \bibinfo{journal}{\emph{ACM Transactions on Graphics (TOG)}} \bibinfo{volume}{39}, \bibinfo{number}{4} (\bibinfo{year}{2020}), \bibinfo{pages}{85--1}.
\newblock


\bibitem[Choi et~al\mbox{.}(2018)]%
        {choi2018stargan}
\bibfield{author}{\bibinfo{person}{Yunjey Choi}, \bibinfo{person}{Minje Choi}, \bibinfo{person}{Munyoung Kim}, \bibinfo{person}{Jung-Woo Ha}, \bibinfo{person}{Sunghun Kim}, {and} \bibinfo{person}{Jaegul Choo}.} \bibinfo{year}{2018}\natexlab{}.
\newblock \showarticletitle{Stargan: Unified generative adversarial networks for multi-domain image-to-image translation}. In \bibinfo{booktitle}{\emph{Proceedings of the IEEE conference on computer vision and pattern recognition}}. \bibinfo{pages}{8789--8797}.
\newblock


\bibitem[Christie and Ellis(1981)]%
        {christie1981photofit}
\bibfield{author}{\bibinfo{person}{Donald~F Christie} {and} \bibinfo{person}{Hadyn~D Ellis}.} \bibinfo{year}{1981}\natexlab{}.
\newblock \showarticletitle{Photofit constructions versus verbal descriptions of faces.}
\newblock \bibinfo{journal}{\emph{Journal of Applied Psychology}} \bibinfo{volume}{66}, \bibinfo{number}{3} (\bibinfo{year}{1981}), \bibinfo{pages}{358}.
\newblock


\bibitem[Cowen et~al\mbox{.}(2014)]%
        {cowen2014neural}
\bibfield{author}{\bibinfo{person}{Alan~S Cowen}, \bibinfo{person}{Marvin~M Chun}, {and} \bibinfo{person}{Brice~A Kuhl}.} \bibinfo{year}{2014}\natexlab{}.
\newblock \showarticletitle{Neural portraits of perception: reconstructing face images from evoked brain activity}.
\newblock \bibinfo{journal}{\emph{Neuroimage}}  \bibinfo{volume}{94} (\bibinfo{year}{2014}), \bibinfo{pages}{12--22}.
\newblock


\bibitem[Dado et~al\mbox{.}(2022)]%
        {dado2022hyperrealistic}
\bibfield{author}{\bibinfo{person}{Thirza Dado}, \bibinfo{person}{Ya{\u{g}}mur G{\"u}{\c{c}}l{\"u}t{\"u}rk}, \bibinfo{person}{Luca Ambrogioni}, \bibinfo{person}{Gabri{\"e}lle Ras}, \bibinfo{person}{Sander Bosch}, \bibinfo{person}{Marcel van Gerven}, {and} \bibinfo{person}{Umut G{\"u}{\c{c}}l{\"u}}.} \bibinfo{year}{2022}\natexlab{}.
\newblock \showarticletitle{Hyperrealistic neural decoding for reconstructing faces from fMRI activations via the GAN latent space}.
\newblock \bibinfo{journal}{\emph{Scientific reports}} \bibinfo{volume}{12}, \bibinfo{number}{1} (\bibinfo{year}{2022}), \bibinfo{pages}{1--9}.
\newblock


\bibitem[Date et~al\mbox{.}(2019)]%
        {date2019deep}
\bibfield{author}{\bibinfo{person}{Hiroto Date}, \bibinfo{person}{Keisuke Kawasaki}, \bibinfo{person}{Isao Hasegawa}, {and} \bibinfo{person}{Takayuki Okatani}.} \bibinfo{year}{2019}\natexlab{}.
\newblock \showarticletitle{Deep learning for natural image reconstruction from electrocorticography signals}. In \bibinfo{booktitle}{\emph{2019 IEEE International Conference on Bioinformatics and Biomedicine}}. \bibinfo{pages}{2331--2336}.
\newblock


\bibitem[Deng et~al\mbox{.}(2019)]%
        {deng2019arcface}
\bibfield{author}{\bibinfo{person}{Jiankang Deng}, \bibinfo{person}{Jia Guo}, \bibinfo{person}{Niannan Xue}, {and} \bibinfo{person}{Stefanos Zafeiriou}.} \bibinfo{year}{2019}\natexlab{}.
\newblock \showarticletitle{Arcface: Additive angular margin loss for deep face recognition}. In \bibinfo{booktitle}{\emph{Proceedings of the IEEE/CVF Conference on Computer Vision and Pattern Recognition}}. \bibinfo{pages}{4690--4699}.
\newblock


\bibitem[Deng et~al\mbox{.}(2020)]%
        {deng2020disentangled}
\bibfield{author}{\bibinfo{person}{Yu Deng}, \bibinfo{person}{Jiaolong Yang}, \bibinfo{person}{Dong Chen}, \bibinfo{person}{Fang Wen}, {and} \bibinfo{person}{Xin Tong}.} \bibinfo{year}{2020}\natexlab{}.
\newblock \showarticletitle{Disentangled and controllable face image generation via 3d imitative-contrastive learning}. In \bibinfo{booktitle}{\emph{Proceedings of the IEEE/CVF conference on computer vision and pattern recognition}}. \bibinfo{pages}{5154--5163}.
\newblock


\bibitem[Devlin et~al\mbox{.}(2018)]%
        {devlin2018bert}
\bibfield{author}{\bibinfo{person}{Jacob Devlin}, \bibinfo{person}{Ming-Wei Chang}, \bibinfo{person}{Kenton Lee}, {and} \bibinfo{person}{Kristina Toutanova}.} \bibinfo{year}{2018}\natexlab{}.
\newblock \showarticletitle{Bert: Pre-training of deep bidirectional transformers for language understanding}.
\newblock \bibinfo{journal}{\emph{arXiv preprint arXiv:1810.04805}} (\bibinfo{year}{2018}).
\newblock


\bibitem[Ellis et~al\mbox{.}(1978)]%
        {ellis1978critical}
\bibfield{author}{\bibinfo{person}{Hadyn~D Ellis}, \bibinfo{person}{Graham~M Davies}, {and} \bibinfo{person}{John~W Shepherd}.} \bibinfo{year}{1978}\natexlab{}.
\newblock \showarticletitle{A critical examination of the Photofit system for recalling faces}.
\newblock \bibinfo{journal}{\emph{Ergonomics}} \bibinfo{volume}{21}, \bibinfo{number}{4} (\bibinfo{year}{1978}), \bibinfo{pages}{297--307}.
\newblock


\bibitem[Farah et~al\mbox{.}(1998)]%
        {farah1998special}
\bibfield{author}{\bibinfo{person}{Martha~J Farah}, \bibinfo{person}{Kevin~D Wilson}, \bibinfo{person}{Maxwell Drain}, {and} \bibinfo{person}{James~N Tanaka}.} \bibinfo{year}{1998}\natexlab{}.
\newblock \showarticletitle{What is" special" about face perception?}
\newblock \bibinfo{journal}{\emph{Psychological Review}} \bibinfo{volume}{105}, \bibinfo{number}{3} (\bibinfo{year}{1998}), \bibinfo{pages}{482}.
\newblock


\bibitem[Frowd et~al\mbox{.}(2004)]%
        {frowd2004evofit}
\bibfield{author}{\bibinfo{person}{Charlie~D Frowd}, \bibinfo{person}{Peter~JB Hancock}, {and} \bibinfo{person}{Derek Carson}.} \bibinfo{year}{2004}\natexlab{}.
\newblock \showarticletitle{EvoFIT: A holistic, evolutionary facial imaging technique for creating composites}.
\newblock \bibinfo{journal}{\emph{ACM Transactions on applied perception}} \bibinfo{volume}{1}, \bibinfo{number}{1} (\bibinfo{year}{2004}), \bibinfo{pages}{19--39}.
\newblock


\bibitem[Gao et~al\mbox{.}(2021)]%
        {gao2021high}
\bibfield{author}{\bibinfo{person}{Yue Gao}, \bibinfo{person}{Fangyun Wei}, \bibinfo{person}{Jianmin Bao}, \bibinfo{person}{Shuyang Gu}, \bibinfo{person}{Dong Chen}, \bibinfo{person}{Fang Wen}, {and} \bibinfo{person}{Zhouhui Lian}.} \bibinfo{year}{2021}\natexlab{}.
\newblock \showarticletitle{High-fidelity and arbitrary face editing}. In \bibinfo{booktitle}{\emph{Proceedings of the IEEE/CVF conference on computer vision and pattern recognition}}. \bibinfo{pages}{16115--16124}.
\newblock


\bibitem[Gibson et~al\mbox{.}(2009)]%
        {gibson2009new}
\bibfield{author}{\bibinfo{person}{Stuart~J Gibson}, \bibinfo{person}{Chris~J Solomon}, \bibinfo{person}{Matthew~IS Maylin}, {and} \bibinfo{person}{Clifford Clark}.} \bibinfo{year}{2009}\natexlab{}.
\newblock \showarticletitle{New methodology in facial composite construction: From theory to practice}.
\newblock \bibinfo{journal}{\emph{International Journal of Electronic Security and Digital Forensics}} \bibinfo{volume}{2}, \bibinfo{number}{2} (\bibinfo{year}{2009}), \bibinfo{pages}{156--168}.
\newblock


\bibitem[Gu et~al\mbox{.}(2019)]%
        {gu2019mask}
\bibfield{author}{\bibinfo{person}{Shuyang Gu}, \bibinfo{person}{Jianmin Bao}, \bibinfo{person}{Hao Yang}, \bibinfo{person}{Dong Chen}, \bibinfo{person}{Fang Wen}, {and} \bibinfo{person}{Lu Yuan}.} \bibinfo{year}{2019}\natexlab{}.
\newblock \showarticletitle{Mask-guided portrait editing with conditional gans}. In \bibinfo{booktitle}{\emph{Proceedings of the IEEE/CVF conference on computer vision and pattern recognition}}. \bibinfo{pages}{3436--3445}.
\newblock


\bibitem[G{\"u}{\c{c}}l{\"u}t{\"u}rk et~al\mbox{.}(2017)]%
        {guccluturk2017reconstructing}
\bibfield{author}{\bibinfo{person}{Ya{\u{g}}mur G{\"u}{\c{c}}l{\"u}t{\"u}rk}, \bibinfo{person}{Umut G{\"u}{\c{c}}l{\"u}}, \bibinfo{person}{Katja Seeliger}, \bibinfo{person}{Sander Bosch}, \bibinfo{person}{Rob van Lier}, {and} \bibinfo{person}{Marcel~A van Gerven}.} \bibinfo{year}{2017}\natexlab{}.
\newblock \showarticletitle{Reconstructing perceived faces from brain activations with deep adversarial neural decoding}.
\newblock \bibinfo{journal}{\emph{Advances in Neural Information Processing Systems}}  \bibinfo{volume}{30} (\bibinfo{year}{2017}).
\newblock


\bibitem[Guo et~al\mbox{.}(2019)]%
        {guo2019mulgan}
\bibfield{author}{\bibinfo{person}{Jingtao Guo}, \bibinfo{person}{Zhenzhen Qian}, \bibinfo{person}{Zuowei Zhou}, {and} \bibinfo{person}{Yi Liu}.} \bibinfo{year}{2019}\natexlab{}.
\newblock \showarticletitle{Mulgan: Facial attribute editing by exemplar}.
\newblock \bibinfo{journal}{\emph{arXiv preprint arXiv:1912.12396}} (\bibinfo{year}{2019}).
\newblock


\bibitem[Han et~al\mbox{.}(2021)]%
        {han2021disentangled}
\bibfield{author}{\bibinfo{person}{Yuxuan Han}, \bibinfo{person}{Jiaolong Yang}, {and} \bibinfo{person}{Ying Fu}.} \bibinfo{year}{2021}\natexlab{}.
\newblock \showarticletitle{Disentangled face attribute editing via instance-aware latent space search}.
\newblock \bibinfo{journal}{\emph{IJCAI}} (\bibinfo{year}{2021}).
\newblock


\bibitem[H{\"a}rk{\"o}nen et~al\mbox{.}(2020)]%
        {harkonen2020ganspace}
\bibfield{author}{\bibinfo{person}{Erik H{\"a}rk{\"o}nen}, \bibinfo{person}{Aaron Hertzmann}, \bibinfo{person}{Jaakko Lehtinen}, {and} \bibinfo{person}{Sylvain Paris}.} \bibinfo{year}{2020}\natexlab{}.
\newblock \showarticletitle{Ganspace: Discovering interpretable gan controls}.
\newblock \bibinfo{journal}{\emph{Advances in neural information processing systems}}  \bibinfo{volume}{33} (\bibinfo{year}{2020}), \bibinfo{pages}{9841--9850}.
\newblock


\bibitem[He et~al\mbox{.}(2016)]%
        {he2016deep}
\bibfield{author}{\bibinfo{person}{Kaiming He}, \bibinfo{person}{Xiangyu Zhang}, \bibinfo{person}{Shaoqing Ren}, {and} \bibinfo{person}{Jian Sun}.} \bibinfo{year}{2016}\natexlab{}.
\newblock \showarticletitle{Deep residual learning for image recognition}. In \bibinfo{booktitle}{\emph{Proceedings of the IEEE Conference on Computer Vision and Pattern Recognition}}. \bibinfo{pages}{770--778}.
\newblock


\bibitem[He et~al\mbox{.}(2020)]%
        {he2020pa}
\bibfield{author}{\bibinfo{person}{Zhenliang He}, \bibinfo{person}{Meina Kan}, \bibinfo{person}{Jichao Zhang}, {and} \bibinfo{person}{Shiguang Shan}.} \bibinfo{year}{2020}\natexlab{}.
\newblock \showarticletitle{Pa-gan: Progressive attention generative adversarial network for facial attribute editing}.
\newblock \bibinfo{journal}{\emph{arXiv preprint arXiv:2007.05892}} (\bibinfo{year}{2020}).
\newblock


\bibitem[He et~al\mbox{.}(2019)]%
        {he2019attgan}
\bibfield{author}{\bibinfo{person}{Zhenliang He}, \bibinfo{person}{Wangmeng Zuo}, \bibinfo{person}{Meina Kan}, \bibinfo{person}{Shiguang Shan}, {and} \bibinfo{person}{Xilin Chen}.} \bibinfo{year}{2019}\natexlab{}.
\newblock \showarticletitle{Attgan: Facial attribute editing by only changing what you want}.
\newblock \bibinfo{journal}{\emph{IEEE transactions on image processing}} \bibinfo{volume}{28}, \bibinfo{number}{11} (\bibinfo{year}{2019}), \bibinfo{pages}{5464--5478}.
\newblock


\bibitem[Hou et~al\mbox{.}(2022a)]%
        {hou2022feat}
\bibfield{author}{\bibinfo{person}{Xianxu Hou}, \bibinfo{person}{Linlin Shen}, \bibinfo{person}{Or Patashnik}, \bibinfo{person}{Daniel Cohen-Or}, {and} \bibinfo{person}{Hui Huang}.} \bibinfo{year}{2022}\natexlab{a}.
\newblock \showarticletitle{Feat: Face editing with attention}.
\newblock \bibinfo{journal}{\emph{arXiv preprint arXiv:2202.02713}} (\bibinfo{year}{2022}).
\newblock


\bibitem[Hou et~al\mbox{.}(2022b)]%
        {hou2022guidedstyle}
\bibfield{author}{\bibinfo{person}{Xianxu Hou}, \bibinfo{person}{Xiaokang Zhang}, \bibinfo{person}{Hanbang Liang}, \bibinfo{person}{Linlin Shen}, \bibinfo{person}{Zhihui Lai}, {and} \bibinfo{person}{Jun Wan}.} \bibinfo{year}{2022}\natexlab{b}.
\newblock \showarticletitle{Guidedstyle: Attribute knowledge guided style manipulation for semantic face editing}.
\newblock \bibinfo{journal}{\emph{Neural Networks}}  \bibinfo{volume}{145} (\bibinfo{year}{2022}), \bibinfo{pages}{209--220}.
\newblock


\bibitem[Huang et~al\mbox{.}(2023)]%
        {huang2023collaborative}
\bibfield{author}{\bibinfo{person}{Ziqi Huang}, \bibinfo{person}{Kelvin~CK Chan}, \bibinfo{person}{Yuming Jiang}, {and} \bibinfo{person}{Ziwei Liu}.} \bibinfo{year}{2023}\natexlab{}.
\newblock \showarticletitle{Collaborative diffusion for multi-modal face generation and editing}. In \bibinfo{booktitle}{\emph{Proceedings of the IEEE/CVF Conference on Computer Vision and Pattern Recognition}}. \bibinfo{pages}{6080--6090}.
\newblock


\bibitem[Ioffe and Szegedy(2015)]%
        {ioffe2015batch}
\bibfield{author}{\bibinfo{person}{Sergey Ioffe} {and} \bibinfo{person}{Christian Szegedy}.} \bibinfo{year}{2015}\natexlab{}.
\newblock \showarticletitle{Batch normalization: Accelerating deep network training by reducing internal covariate shift}. In \bibinfo{booktitle}{\emph{International Conference on Machine Learning}}. \bibinfo{pages}{448--456}.
\newblock


\bibitem[Jeannerod(1995)]%
        {jeannerod1995mental}
\bibfield{author}{\bibinfo{person}{Marc Jeannerod}.} \bibinfo{year}{1995}\natexlab{}.
\newblock \showarticletitle{Mental imagery in the motor context}.
\newblock \bibinfo{journal}{\emph{Neuropsychologia}} \bibinfo{volume}{33}, \bibinfo{number}{11} (\bibinfo{year}{1995}), \bibinfo{pages}{1419--1432}.
\newblock


\bibitem[Karras et~al\mbox{.}(2017)]%
        {karras2017progressive}
\bibfield{author}{\bibinfo{person}{Tero Karras}, \bibinfo{person}{Timo Aila}, \bibinfo{person}{Samuli Laine}, {and} \bibinfo{person}{Jaakko Lehtinen}.} \bibinfo{year}{2017}\natexlab{}.
\newblock \showarticletitle{Progressive growing of gans for improved quality, stability, and variation}.
\newblock \bibinfo{journal}{\emph{arXiv preprint arXiv:1710.10196}} (\bibinfo{year}{2017}).
\newblock


\bibitem[Karras et~al\mbox{.}(2019)]%
        {karras2019style}
\bibfield{author}{\bibinfo{person}{Tero Karras}, \bibinfo{person}{Samuli Laine}, {and} \bibinfo{person}{Timo Aila}.} \bibinfo{year}{2019}\natexlab{}.
\newblock \showarticletitle{A style-based generator architecture for generative adversarial networks}. In \bibinfo{booktitle}{\emph{Proceedings of the IEEE/CVF Conference on Computer Vision and Pattern Recognition}}. \bibinfo{pages}{4401--4410}.
\newblock


\bibitem[Karras et~al\mbox{.}(2020)]%
        {karras2020analyzing}
\bibfield{author}{\bibinfo{person}{Tero Karras}, \bibinfo{person}{Samuli Laine}, \bibinfo{person}{Miika Aittala}, \bibinfo{person}{Janne Hellsten}, \bibinfo{person}{Jaakko Lehtinen}, {and} \bibinfo{person}{Timo Aila}.} \bibinfo{year}{2020}\natexlab{}.
\newblock \showarticletitle{Analyzing and improving the image quality of stylegan}. In \bibinfo{booktitle}{\emph{Proceedings of the IEEE/CVF Conference on Computer Vision and Pattern Recognition}}. \bibinfo{pages}{8110--8119}.
\newblock


\bibitem[Khodadadeh et~al\mbox{.}(2022)]%
        {khodadadeh2022latent}
\bibfield{author}{\bibinfo{person}{Siavash Khodadadeh}, \bibinfo{person}{Shabnam Ghadar}, \bibinfo{person}{Saeid Motiian}, \bibinfo{person}{Wei-An Lin}, \bibinfo{person}{Ladislau B{\"o}l{\"o}ni}, {and} \bibinfo{person}{Ratheesh Kalarot}.} \bibinfo{year}{2022}\natexlab{}.
\newblock \showarticletitle{Latent to latent: A learned mapper for identity preserving editing of multiple face attributes in stylegan-generated images}. In \bibinfo{booktitle}{\emph{Proceedings of the IEEE/CVF Winter Conference on Applications of Computer Vision}}. \bibinfo{pages}{3184--3192}.
\newblock


\bibitem[Kim et~al\mbox{.}(2021)]%
        {kim2021roles}
\bibfield{author}{\bibinfo{person}{Minjeong Kim}, \bibinfo{person}{Jung-Hwan Kim}, \bibinfo{person}{Minjung Park}, {and} \bibinfo{person}{Jungmin Yoo}.} \bibinfo{year}{2021}\natexlab{}.
\newblock \showarticletitle{The roles of sensory perceptions and mental imagery in consumer decision-making}.
\newblock \bibinfo{journal}{\emph{Journal of Retailing and Consumer Services}}  \bibinfo{volume}{61} (\bibinfo{year}{2021}), \bibinfo{pages}{102517}.
\newblock


\bibitem[Kingma and Ba(2014)]%
        {kingma2014adam}
\bibfield{author}{\bibinfo{person}{Diederik~P Kingma} {and} \bibinfo{person}{Jimmy Ba}.} \bibinfo{year}{2014}\natexlab{}.
\newblock \showarticletitle{Adam: A method for stochastic optimization}.
\newblock \bibinfo{journal}{\emph{arXiv preprint arXiv:1412.6980}} (\bibinfo{year}{2014}).
\newblock


\bibitem[Koehn and Fisher(1997)]%
        {koehn1997constructing}
\bibfield{author}{\bibinfo{person}{Christine~E Koehn} {and} \bibinfo{person}{Ronald~P Fisher}.} \bibinfo{year}{1997}\natexlab{}.
\newblock \showarticletitle{Constructing facial composites with the Mac-a-Mug Pro system}.
\newblock \bibinfo{journal}{\emph{Psychology, Crime and Law}} \bibinfo{volume}{3}, \bibinfo{number}{3} (\bibinfo{year}{1997}), \bibinfo{pages}{209--218}.
\newblock


\bibitem[Kowalski et~al\mbox{.}(2020)]%
        {kowalski2020config}
\bibfield{author}{\bibinfo{person}{Marek Kowalski}, \bibinfo{person}{Stephan~J Garbin}, \bibinfo{person}{Virginia Estellers}, \bibinfo{person}{Tadas Baltru{\v{s}}aitis}, \bibinfo{person}{Matthew Johnson}, {and} \bibinfo{person}{Jamie Shotton}.} \bibinfo{year}{2020}\natexlab{}.
\newblock \showarticletitle{Config: Controllable neural face image generation}. In \bibinfo{booktitle}{\emph{Computer Vision--ECCV 2020: 16th European Conference, Glasgow, UK, August 23--28, 2020, Proceedings, Part XI 16}}. Springer, \bibinfo{pages}{299--315}.
\newblock


\bibitem[Kwak et~al\mbox{.}(2020)]%
        {kwak2020cafe}
\bibfield{author}{\bibinfo{person}{Jeong-gi Kwak}, \bibinfo{person}{David~K Han}, {and} \bibinfo{person}{Hanseok Ko}.} \bibinfo{year}{2020}\natexlab{}.
\newblock \showarticletitle{CAFE-GAN: Arbitrary face attribute editing with complementary attention feature}. In \bibinfo{booktitle}{\emph{Computer Vision--ECCV 2020: 16th European Conference, Glasgow, UK, August 23--28, 2020, Proceedings, Part XIV 16}}. Springer, \bibinfo{pages}{524--540}.
\newblock


\bibitem[Laughery and Fowler(1980)]%
        {Laughery1980}
\bibfield{author}{\bibinfo{person}{Kenneth~R. Laughery} {and} \bibinfo{person}{Richard~H. Fowler}.} \bibinfo{year}{1980}\natexlab{}.
\newblock \showarticletitle{Sketch artist and Identi-kit procedures for recalling faces.}
\newblock \bibinfo{journal}{\emph{Journal of Applied Psychology}} \bibinfo{volume}{65}, \bibinfo{number}{3} (\bibinfo{date}{June} \bibinfo{year}{1980}), \bibinfo{pages}{307--316}.
\newblock


\bibitem[Lee et~al\mbox{.}(2020)]%
        {lee2020maskgan}
\bibfield{author}{\bibinfo{person}{Cheng-Han Lee}, \bibinfo{person}{Ziwei Liu}, \bibinfo{person}{Lingyun Wu}, {and} \bibinfo{person}{Ping Luo}.} \bibinfo{year}{2020}\natexlab{}.
\newblock \showarticletitle{Maskgan: Towards diverse and interactive facial image manipulation}. In \bibinfo{booktitle}{\emph{Proceedings of the IEEE/CVF Conference on Computer Vision and Pattern Recognition}}. \bibinfo{pages}{5549--5558}.
\newblock


\bibitem[Lin et~al\mbox{.}(2019)]%
        {lin2019dcnn}
\bibfield{author}{\bibinfo{person}{Yunfeng Lin}, \bibinfo{person}{Jiangbei Li}, {and} \bibinfo{person}{Hanjing Wang}.} \bibinfo{year}{2019}\natexlab{}.
\newblock \showarticletitle{DCNN-GAN: Reconstructing Realistic Image from fMRI}. In \bibinfo{booktitle}{\emph{2019 16th International Conference on Machine Vision Applications}}. \bibinfo{pages}{1--6}.
\newblock


\bibitem[Ling et~al\mbox{.}(2021)]%
        {ling2021editgan}
\bibfield{author}{\bibinfo{person}{Huan Ling}, \bibinfo{person}{Karsten Kreis}, \bibinfo{person}{Daiqing Li}, \bibinfo{person}{Seung~Wook Kim}, \bibinfo{person}{Antonio Torralba}, {and} \bibinfo{person}{Sanja Fidler}.} \bibinfo{year}{2021}\natexlab{}.
\newblock \showarticletitle{Editgan: High-precision semantic image editing}.
\newblock \bibinfo{journal}{\emph{Advances in Neural Information Processing Systems}}  \bibinfo{volume}{34} (\bibinfo{year}{2021}), \bibinfo{pages}{16331--16345}.
\newblock


\bibitem[Lu et~al\mbox{.}(2018)]%
        {lu2018attribute}
\bibfield{author}{\bibinfo{person}{Yongyi Lu}, \bibinfo{person}{Yu-Wing Tai}, {and} \bibinfo{person}{Chi-Keung Tang}.} \bibinfo{year}{2018}\natexlab{}.
\newblock \showarticletitle{Attribute-guided face generation using conditional cyclegan}. In \bibinfo{booktitle}{\emph{Proceedings of the European conference on computer vision (ECCV)}}. \bibinfo{pages}{282--297}.
\newblock


\bibitem[Medin et~al\mbox{.}(2022)]%
        {medin2022most}
\bibfield{author}{\bibinfo{person}{Safa~C Medin}, \bibinfo{person}{Bernhard Egger}, \bibinfo{person}{Anoop Cherian}, \bibinfo{person}{Ye Wang}, \bibinfo{person}{Joshua~B Tenenbaum}, \bibinfo{person}{Xiaoming Liu}, {and} \bibinfo{person}{Tim~K Marks}.} \bibinfo{year}{2022}\natexlab{}.
\newblock \showarticletitle{MOST-GAN: 3D morphable StyleGAN for disentangled face image manipulation}. In \bibinfo{booktitle}{\emph{Proceedings of the AAAI conference on artificial intelligence}}, Vol.~\bibinfo{volume}{36}. \bibinfo{pages}{1962--1971}.
\newblock


\bibitem[Moulton and Kosslyn(2009)]%
        {moulton2009imagining}
\bibfield{author}{\bibinfo{person}{Samuel~T Moulton} {and} \bibinfo{person}{Stephen~M Kosslyn}.} \bibinfo{year}{2009}\natexlab{}.
\newblock \showarticletitle{Imagining predictions: mental imagery as mental emulation}.
\newblock \bibinfo{journal}{\emph{Philosophical Transactions of the Royal Society B: Biological Sciences}} \bibinfo{volume}{364}, \bibinfo{number}{1521} (\bibinfo{year}{2009}), \bibinfo{pages}{1273--1280}.
\newblock


\bibitem[Naselaris et~al\mbox{.}(2015)]%
        {naselaris2015voxel}
\bibfield{author}{\bibinfo{person}{Thomas Naselaris}, \bibinfo{person}{Cheryl~A Olman}, \bibinfo{person}{Dustin~E Stansbury}, \bibinfo{person}{Kamil Ugurbil}, {and} \bibinfo{person}{Jack~L Gallant}.} \bibinfo{year}{2015}\natexlab{}.
\newblock \showarticletitle{A voxel-wise encoding model for early visual areas decodes mental images of remembered scenes}.
\newblock \bibinfo{journal}{\emph{Neuroimage}}  \bibinfo{volume}{105} (\bibinfo{year}{2015}), \bibinfo{pages}{215--228}.
\newblock


\bibitem[Nemrodov et~al\mbox{.}(2018)]%
        {nemrodov2018neural}
\bibfield{author}{\bibinfo{person}{Dan Nemrodov}, \bibinfo{person}{Matthias Niemeier}, \bibinfo{person}{Ashutosh Patel}, {and} \bibinfo{person}{Adrian Nestor}.} \bibinfo{year}{2018}\natexlab{}.
\newblock \showarticletitle{The neural dynamics of facial identity processing: insights from EEG-based pattern analysis and image reconstruction}.
\newblock \bibinfo{journal}{\emph{Eneuro}} \bibinfo{volume}{5}, \bibinfo{number}{1} (\bibinfo{year}{2018}).
\newblock


\bibitem[Nestor et~al\mbox{.}(2016)]%
        {nestor2016feature}
\bibfield{author}{\bibinfo{person}{Adrian Nestor}, \bibinfo{person}{David~C Plaut}, {and} \bibinfo{person}{Marlene Behrmann}.} \bibinfo{year}{2016}\natexlab{}.
\newblock \showarticletitle{Feature-based face representations and image reconstruction from behavioral and neural data}.
\newblock \bibinfo{journal}{\emph{Proceedings of the National Academy of Sciences}} \bibinfo{volume}{113}, \bibinfo{number}{2} (\bibinfo{year}{2016}), \bibinfo{pages}{416--421}.
\newblock


\bibitem[Niu et~al\mbox{.}(2023)]%
        {niu2023disentangling}
\bibfield{author}{\bibinfo{person}{Yongjie Niu}, \bibinfo{person}{Mingquan Zhou}, {and} \bibinfo{person}{Zhan Li}.} \bibinfo{year}{2023}\natexlab{}.
\newblock \showarticletitle{Disentangling the latent space of GANs for semantic face editing}.
\newblock \bibinfo{journal}{\emph{Plos one}} \bibinfo{volume}{18}, \bibinfo{number}{10} (\bibinfo{year}{2023}), \bibinfo{pages}{e0293496}.
\newblock


\bibitem[Ozcelik and VanRullen(2023)]%
        {ozcelik2023natural}
\bibfield{author}{\bibinfo{person}{Furkan Ozcelik} {and} \bibinfo{person}{Rufin VanRullen}.} \bibinfo{year}{2023}\natexlab{}.
\newblock \showarticletitle{Natural scene reconstruction from fMRI signals using generative latent diffusion}.
\newblock \bibinfo{journal}{\emph{Scientific Reports}} \bibinfo{volume}{13}, \bibinfo{number}{1} (\bibinfo{year}{2023}), \bibinfo{pages}{15666}.
\newblock


\bibitem[Pan et~al\mbox{.}(2023)]%
        {pan2023drag}
\bibfield{author}{\bibinfo{person}{Xingang Pan}, \bibinfo{person}{Ayush Tewari}, \bibinfo{person}{Thomas Leimk{\"u}hler}, \bibinfo{person}{Lingjie Liu}, \bibinfo{person}{Abhimitra Meka}, {and} \bibinfo{person}{Christian Theobalt}.} \bibinfo{year}{2023}\natexlab{}.
\newblock \showarticletitle{Drag your gan: Interactive point-based manipulation on the generative image manifold}. In \bibinfo{booktitle}{\emph{ACM SIGGRAPH 2023 Conference Proceedings}}. \bibinfo{pages}{1--11}.
\newblock


\bibitem[Parkhi et~al\mbox{.}(2015)]%
        {parkhi2015deep}
\bibfield{author}{\bibinfo{person}{Omkar~M Parkhi}, \bibinfo{person}{Andrea Vedaldi}, {and} \bibinfo{person}{Andrew Zisserman}.} \bibinfo{year}{2015}\natexlab{}.
\newblock \showarticletitle{Deep face recognition}. In \bibinfo{booktitle}{\emph{Proceedings of the British Machine Vision Conference}}.
\newblock


\bibitem[Patashnik et~al\mbox{.}(2021)]%
        {patashnik2021styleclip}
\bibfield{author}{\bibinfo{person}{Or Patashnik}, \bibinfo{person}{Zongze Wu}, \bibinfo{person}{Eli Shechtman}, \bibinfo{person}{Daniel Cohen-Or}, {and} \bibinfo{person}{Dani Lischinski}.} \bibinfo{year}{2021}\natexlab{}.
\newblock \showarticletitle{Styleclip: Text-driven manipulation of stylegan imagery}. In \bibinfo{booktitle}{\emph{Proceedings of the IEEE/CVF International Conference on Computer Vision}}. \bibinfo{pages}{2085--2094}.
\newblock


\bibitem[Pearson et~al\mbox{.}(2015)]%
        {pearson2015mental}
\bibfield{author}{\bibinfo{person}{Joel Pearson}, \bibinfo{person}{Thomas Naselaris}, \bibinfo{person}{Emily~A Holmes}, {and} \bibinfo{person}{Stephen~M Kosslyn}.} \bibinfo{year}{2015}\natexlab{}.
\newblock \showarticletitle{Mental imagery: functional mechanisms and clinical applications}.
\newblock \bibinfo{journal}{\emph{Trends in cognitive sciences}} \bibinfo{volume}{19}, \bibinfo{number}{10} (\bibinfo{year}{2015}), \bibinfo{pages}{590--602}.
\newblock


\bibitem[Portenier et~al\mbox{.}(2018)]%
        {portenier2018faceshop}
\bibfield{author}{\bibinfo{person}{Tiziano Portenier}, \bibinfo{person}{Qiyang Hu}, \bibinfo{person}{Attila Szab{\'o}}, \bibinfo{person}{Siavash~Arjomand Bigdeli}, \bibinfo{person}{Paolo Favaro}, {and} \bibinfo{person}{Matthias Zwicker}.} \bibinfo{year}{2018}\natexlab{}.
\newblock \showarticletitle{Faceshop: deep sketch-based face image editing}.
\newblock \bibinfo{journal}{\emph{ACM Transactions on Graphics (TOG)}} \bibinfo{volume}{37}, \bibinfo{number}{4} (\bibinfo{year}{2018}), \bibinfo{pages}{1--13}.
\newblock


\bibitem[Sadovnik et~al\mbox{.}(2018)]%
        {sadovnik2018finding}
\bibfield{author}{\bibinfo{person}{Amir Sadovnik}, \bibinfo{person}{Wassim Gharbi}, \bibinfo{person}{Thanh Vu}, {and} \bibinfo{person}{Andrew Gallagher}.} \bibinfo{year}{2018}\natexlab{}.
\newblock \showarticletitle{Finding your lookalike: Measuring face similarity rather than face identity}. In \bibinfo{booktitle}{\emph{Proceedings of the IEEE Conference on Computer Vision and Pattern Recognition Workshops}}. \bibinfo{pages}{2345--2353}.
\newblock


\bibitem[Saravanos et~al\mbox{.}(2021)]%
        {saravanos2021hidden}
\bibfield{author}{\bibinfo{person}{Antonios Saravanos}, \bibinfo{person}{Stavros Zervoudakis}, \bibinfo{person}{Dongnanzi Zheng}, \bibinfo{person}{Neil Stott}, \bibinfo{person}{Bohdan Hawryluk}, {and} \bibinfo{person}{Donatella Delfino}.} \bibinfo{year}{2021}\natexlab{}.
\newblock \showarticletitle{The hidden cost of using Amazon Mechanical Turk for research}. In \bibinfo{booktitle}{\emph{HCI International 2021-Late Breaking Papers: Design and User Experience: 23rd HCI International Conference, HCII 2021, Virtual Event, July 24--29, 2021, Proceedings 23}}. Springer, \bibinfo{pages}{147--164}.
\newblock


\bibitem[Sattar et~al\mbox{.}(2017)]%
        {sattar2017predicting}
\bibfield{author}{\bibinfo{person}{Hosnieh Sattar}, \bibinfo{person}{Andreas Bulling}, {and} \bibinfo{person}{Mario Fritz}.} \bibinfo{year}{2017}\natexlab{}.
\newblock \showarticletitle{Predicting the category and attributes of visual search targets using deep gaze pooling}. In \bibinfo{booktitle}{\emph{Proceedings of the IEEE International Conference on Computer Vision Workshops}}. \bibinfo{pages}{2740--2748}.
\newblock


\bibitem[Sattar et~al\mbox{.}(2020)]%
        {sattar2020deep}
\bibfield{author}{\bibinfo{person}{Hosnieh Sattar}, \bibinfo{person}{Mario Fritz}, {and} \bibinfo{person}{Andreas Bulling}.} \bibinfo{year}{2020}\natexlab{}.
\newblock \showarticletitle{Deep gaze pooling: Inferring and visually decoding search intents from human gaze fixations}.
\newblock \bibinfo{journal}{\emph{Neurocomputing}}  \bibinfo{volume}{387} (\bibinfo{year}{2020}), \bibinfo{pages}{369--382}.
\newblock


\bibitem[Scotti et~al\mbox{.}(2024)]%
        {scotti2024reconstructing}
\bibfield{author}{\bibinfo{person}{Paul Scotti}, \bibinfo{person}{Atmadeep Banerjee}, \bibinfo{person}{Jimmie Goode}, \bibinfo{person}{Stepan Shabalin}, \bibinfo{person}{Alex Nguyen}, \bibinfo{person}{Aidan Dempster}, \bibinfo{person}{Nathalie Verlinde}, \bibinfo{person}{Elad Yundler}, \bibinfo{person}{David Weisberg}, \bibinfo{person}{Kenneth Norman}, {et~al\mbox{.}}} \bibinfo{year}{2024}\natexlab{}.
\newblock \showarticletitle{Reconstructing the mind's eye: fMRI-to-image with contrastive learning and diffusion priors}.
\newblock \bibinfo{journal}{\emph{Advances in Neural Information Processing Systems}}  \bibinfo{volume}{36} (\bibinfo{year}{2024}).
\newblock


\bibitem[Seeliger et~al\mbox{.}(2018)]%
        {seeliger2018generative}
\bibfield{author}{\bibinfo{person}{Katja Seeliger}, \bibinfo{person}{Umut G{\"u}{\c{c}}l{\"u}}, \bibinfo{person}{Luca Ambrogioni}, \bibinfo{person}{Yagmur G{\"u}{\c{c}}l{\"u}t{\"u}rk}, {and} \bibinfo{person}{Marcel~AJ van Gerven}.} \bibinfo{year}{2018}\natexlab{}.
\newblock \showarticletitle{Generative adversarial networks for reconstructing natural images from brain activity}.
\newblock \bibinfo{journal}{\emph{NeuroImage}}  \bibinfo{volume}{181} (\bibinfo{year}{2018}), \bibinfo{pages}{775--785}.
\newblock


\bibitem[Shatek et~al\mbox{.}(2019)]%
        {shatek2019decoding}
\bibfield{author}{\bibinfo{person}{Sophia~M Shatek}, \bibinfo{person}{Tijl Grootswagers}, \bibinfo{person}{Amanda~K Robinson}, {and} \bibinfo{person}{Thomas~A Carlson}.} \bibinfo{year}{2019}\natexlab{}.
\newblock \showarticletitle{Decoding images in the mind’s eye: The temporal dynamics of visual imagery}.
\newblock \bibinfo{journal}{\emph{Vision}} \bibinfo{volume}{3}, \bibinfo{number}{4} (\bibinfo{year}{2019}), \bibinfo{pages}{53}.
\newblock


\bibitem[Shen et~al\mbox{.}(2019)]%
        {shen2019deep}
\bibfield{author}{\bibinfo{person}{Guohua Shen}, \bibinfo{person}{Tomoyasu Horikawa}, \bibinfo{person}{Kei Majima}, {and} \bibinfo{person}{Yukiyasu Kamitani}.} \bibinfo{year}{2019}\natexlab{}.
\newblock \showarticletitle{Deep image reconstruction from human brain activity}.
\newblock \bibinfo{journal}{\emph{PLoS computational biology}} \bibinfo{volume}{15}, \bibinfo{number}{1} (\bibinfo{year}{2019}), \bibinfo{pages}{e1006633}.
\newblock


\bibitem[Shen et~al\mbox{.}(2020)]%
        {shen2020interfacegan}
\bibfield{author}{\bibinfo{person}{Yujun Shen}, \bibinfo{person}{Ceyuan Yang}, \bibinfo{person}{Xiaoou Tang}, {and} \bibinfo{person}{Bolei Zhou}.} \bibinfo{year}{2020}\natexlab{}.
\newblock \showarticletitle{Interfacegan: Interpreting the disentangled face representation learned by gans}.
\newblock \bibinfo{journal}{\emph{IEEE transactions on pattern analysis and machine intelligence}} \bibinfo{volume}{44}, \bibinfo{number}{4} (\bibinfo{year}{2020}), \bibinfo{pages}{2004--2018}.
\newblock


\bibitem[Shen and Zhou(2021)]%
        {shen2021closed}
\bibfield{author}{\bibinfo{person}{Yujun Shen} {and} \bibinfo{person}{Bolei Zhou}.} \bibinfo{year}{2021}\natexlab{}.
\newblock \showarticletitle{Closed-form factorization of latent semantics in gans}. In \bibinfo{booktitle}{\emph{Proceedings of the IEEE/CVF conference on computer vision and pattern recognition}}. \bibinfo{pages}{1532--1540}.
\newblock


\bibitem[Sims and Missal(2019)]%
        {sims2019perceptual}
\bibfield{author}{\bibinfo{person}{Andrew Sims} {and} \bibinfo{person}{Marcus Missal}.} \bibinfo{year}{2019}\natexlab{}.
\newblock \showarticletitle{Perceptual Decision-Making and Beyond: Intention as Mental Imagery}.
\newblock In \bibinfo{booktitle}{\emph{Free Will, Causality, and Neuroscience}}. \bibinfo{publisher}{Brill}, \bibinfo{pages}{13--34}.
\newblock


\bibitem[Sinha et~al\mbox{.}(2006)]%
        {sinha2006face}
\bibfield{author}{\bibinfo{person}{Pawan Sinha}, \bibinfo{person}{Benjamin Balas}, \bibinfo{person}{Yuri Ostrovsky}, {and} \bibinfo{person}{Richard Russell}.} \bibinfo{year}{2006}\natexlab{}.
\newblock \showarticletitle{Face recognition by humans: Nineteen results all computer vision researchers should know about}.
\newblock \bibinfo{journal}{\emph{Proc. IEEE}} \bibinfo{volume}{94}, \bibinfo{number}{11} (\bibinfo{year}{2006}), \bibinfo{pages}{1948--1962}.
\newblock


\bibitem[Song et~al\mbox{.}(2019)]%
        {song2019geometry}
\bibfield{author}{\bibinfo{person}{Linsen Song}, \bibinfo{person}{Jie Cao}, \bibinfo{person}{Lingxiao Song}, \bibinfo{person}{Yibo Hu}, {and} \bibinfo{person}{Ran He}.} \bibinfo{year}{2019}\natexlab{}.
\newblock \showarticletitle{Geometry-aware face completion and editing}. In \bibinfo{booktitle}{\emph{Proceedings of the AAAI conference on artificial intelligence}}, Vol.~\bibinfo{volume}{33}. \bibinfo{pages}{2506--2513}.
\newblock


\bibitem[Srivastava et~al\mbox{.}(2014)]%
        {srivastava2014dropout}
\bibfield{author}{\bibinfo{person}{Nitish Srivastava}, \bibinfo{person}{Geoffrey Hinton}, \bibinfo{person}{Alex Krizhevsky}, \bibinfo{person}{Ilya Sutskever}, {and} \bibinfo{person}{Ruslan Salakhutdinov}.} \bibinfo{year}{2014}\natexlab{}.
\newblock \showarticletitle{Dropout: a simple way to prevent neural networks from overfitting}.
\newblock \bibinfo{journal}{\emph{The Journal of Machine Learning Research}} \bibinfo{volume}{15}, \bibinfo{number}{1} (\bibinfo{year}{2014}), \bibinfo{pages}{1929--1958}.
\newblock


\bibitem[Strohm et~al\mbox{.}(2023)]%
        {strohm2023usable}
\bibfield{author}{\bibinfo{person}{Florian Strohm}, \bibinfo{person}{Mihai B{\^a}ce}, {and} \bibinfo{person}{Andreas Bulling}.} \bibinfo{year}{2023}\natexlab{}.
\newblock \showarticletitle{Usable and fast interactive mental face reconstruction}. In \bibinfo{booktitle}{\emph{Proceedings of the 36th Annual ACM Symposium on User Interface Software and Technology}}. \bibinfo{pages}{1--15}.
\newblock


\bibitem[Strohm et~al\mbox{.}(2024)]%
        {strohm2024upface}
\bibfield{author}{\bibinfo{person}{Florian Strohm}, \bibinfo{person}{Mihai B{\^a}ce}, \bibinfo{person}{Markus Kaltenecker}, {and} \bibinfo{person}{Andreas Bulling}.} \bibinfo{year}{2024}\natexlab{}.
\newblock \showarticletitle{SeFFeC: Semantic Facial Feature Control for Fine-grained Face Editing}.
\newblock \bibinfo{journal}{\emph{arXiv preprint arXiv:2403.13972}} (\bibinfo{year}{2024}).
\newblock


\bibitem[Strohm et~al\mbox{.}(2021)]%
        {strohm2021neural}
\bibfield{author}{\bibinfo{person}{Florian Strohm}, \bibinfo{person}{Ekta Sood}, \bibinfo{person}{Sven Mayer}, \bibinfo{person}{Philipp M{\"u}ller}, \bibinfo{person}{Mihai B{\^a}ce}, {and} \bibinfo{person}{Andreas Bulling}.} \bibinfo{year}{2021}\natexlab{}.
\newblock \showarticletitle{Neural Photofit: Gaze-based Mental Image Reconstruction}. In \bibinfo{booktitle}{\emph{Proceedings of the IEEE/CVF International Conference on Computer Vision}}. \bibinfo{pages}{245--254}.
\newblock


\bibitem[Strohm et~al\mbox{.}(2022)]%
        {strohm2022facial}
\bibfield{author}{\bibinfo{person}{Florian Strohm}, \bibinfo{person}{Ekta Sood}, \bibinfo{person}{Dominike Thomas}, \bibinfo{person}{Mihai B{\^a}ce}, {and} \bibinfo{person}{Andreas Bulling}.} \bibinfo{year}{2022}\natexlab{}.
\newblock \showarticletitle{Facial Composite Generation with Iterative Human Feedback}. In \bibinfo{booktitle}{\emph{Proceedings of Machine Learning Research}}.
\newblock


\bibitem[Sun et~al\mbox{.}(2022a)]%
        {sun2022anyface}
\bibfield{author}{\bibinfo{person}{Jianxin Sun}, \bibinfo{person}{Qiyao Deng}, \bibinfo{person}{Qi Li}, \bibinfo{person}{Muyi Sun}, \bibinfo{person}{Min Ren}, {and} \bibinfo{person}{Zhenan Sun}.} \bibinfo{year}{2022}\natexlab{a}.
\newblock \showarticletitle{Anyface: Free-style text-to-face synthesis and manipulation}. In \bibinfo{booktitle}{\emph{Proceedings of the IEEE/CVF Conference on Computer Vision and Pattern Recognition}}. \bibinfo{pages}{18687--18696}.
\newblock


\bibitem[Sun et~al\mbox{.}(2022c)]%
        {sun2022ide}
\bibfield{author}{\bibinfo{person}{Jingxiang Sun}, \bibinfo{person}{Xuan Wang}, \bibinfo{person}{Yichun Shi}, \bibinfo{person}{Lizhen Wang}, \bibinfo{person}{Jue Wang}, {and} \bibinfo{person}{Yebin Liu}.} \bibinfo{year}{2022}\natexlab{c}.
\newblock \showarticletitle{Ide-3d: Interactive disentangled editing for high-resolution 3d-aware portrait synthesis}.
\newblock \bibinfo{journal}{\emph{ACM Transactions on Graphics (ToG)}} \bibinfo{volume}{41}, \bibinfo{number}{6} (\bibinfo{year}{2022}), \bibinfo{pages}{1--10}.
\newblock


\bibitem[Sun et~al\mbox{.}(2022d)]%
        {sun2022fenerf}
\bibfield{author}{\bibinfo{person}{Jingxiang Sun}, \bibinfo{person}{Xuan Wang}, \bibinfo{person}{Yong Zhang}, \bibinfo{person}{Xiaoyu Li}, \bibinfo{person}{Qi Zhang}, \bibinfo{person}{Yebin Liu}, {and} \bibinfo{person}{Jue Wang}.} \bibinfo{year}{2022}\natexlab{d}.
\newblock \showarticletitle{Fenerf: Face editing in neural radiance fields}. In \bibinfo{booktitle}{\emph{Proceedings of the IEEE/CVF Conference on Computer Vision and Pattern Recognition}}. \bibinfo{pages}{7672--7682}.
\newblock


\bibitem[Sun et~al\mbox{.}(2022b)]%
        {sun2022pattgan}
\bibfield{author}{\bibinfo{person}{Qiushi Sun}, \bibinfo{person}{Jingtao Guo}, {and} \bibinfo{person}{Yi Liu}.} \bibinfo{year}{2022}\natexlab{b}.
\newblock \showarticletitle{PattGAN: Pluralistic Facial Attribute Editing}.
\newblock \bibinfo{journal}{\emph{IEEE Access}}  \bibinfo{volume}{10} (\bibinfo{year}{2022}), \bibinfo{pages}{68534--68544}.
\newblock


\bibitem[Takagi and Nishimoto(2023)]%
        {takagi2023high}
\bibfield{author}{\bibinfo{person}{Yu Takagi} {and} \bibinfo{person}{Shinji Nishimoto}.} \bibinfo{year}{2023}\natexlab{}.
\newblock \showarticletitle{High-resolution image reconstruction with latent diffusion models from human brain activity}. In \bibinfo{booktitle}{\emph{Proceedings of the IEEE/CVF Conference on Computer Vision and Pattern Recognition}}. \bibinfo{pages}{14453--14463}.
\newblock


\bibitem[Tewari et~al\mbox{.}(2020a)]%
        {tewari2020pie}
\bibfield{author}{\bibinfo{person}{Ayush Tewari}, \bibinfo{person}{Mohamed Elgharib}, \bibinfo{person}{Florian Bernard}, \bibinfo{person}{Hans-Peter Seidel}, \bibinfo{person}{Patrick P{\'e}rez}, \bibinfo{person}{Michael Zollh{\"o}fer}, {and} \bibinfo{person}{Christian Theobalt}.} \bibinfo{year}{2020}\natexlab{a}.
\newblock \showarticletitle{Pie: Portrait image embedding for semantic control}.
\newblock \bibinfo{journal}{\emph{ACM Transactions on Graphics (TOG)}} \bibinfo{volume}{39}, \bibinfo{number}{6} (\bibinfo{year}{2020}), \bibinfo{pages}{1--14}.
\newblock


\bibitem[Tewari et~al\mbox{.}(2020b)]%
        {tewari2020stylerig}
\bibfield{author}{\bibinfo{person}{Ayush Tewari}, \bibinfo{person}{Mohamed Elgharib}, \bibinfo{person}{Gaurav Bharaj}, \bibinfo{person}{Florian Bernard}, \bibinfo{person}{Hans-Peter Seidel}, \bibinfo{person}{Patrick P{\'e}rez}, \bibinfo{person}{Michael Zollhofer}, {and} \bibinfo{person}{Christian Theobalt}.} \bibinfo{year}{2020}\natexlab{b}.
\newblock \showarticletitle{Stylerig: Rigging stylegan for 3d control over portrait images}. In \bibinfo{booktitle}{\emph{Proceedings of the IEEE/CVF Conference on Computer Vision and Pattern Recognition}}. \bibinfo{pages}{6142--6151}.
\newblock


\bibitem[VanRullen and Reddy(2019)]%
        {vanrullen2019reconstructing}
\bibfield{author}{\bibinfo{person}{Rufin VanRullen} {and} \bibinfo{person}{Leila Reddy}.} \bibinfo{year}{2019}\natexlab{}.
\newblock \showarticletitle{Reconstructing faces from fMRI patterns using deep generative neural networks}.
\newblock \bibinfo{journal}{\emph{Communications biology}} \bibinfo{volume}{2}, \bibinfo{number}{1} (\bibinfo{year}{2019}), \bibinfo{pages}{1--10}.
\newblock


\bibitem[Vaswani et~al\mbox{.}(2017)]%
        {vaswani2017attention}
\bibfield{author}{\bibinfo{person}{Ashish Vaswani}, \bibinfo{person}{Noam Shazeer}, \bibinfo{person}{Niki Parmar}, \bibinfo{person}{Jakob Uszkoreit}, \bibinfo{person}{Llion Jones}, \bibinfo{person}{Aidan~N Gomez}, \bibinfo{person}{{\L}ukasz Kaiser}, {and} \bibinfo{person}{Illia Polosukhin}.} \bibinfo{year}{2017}\natexlab{}.
\newblock \showarticletitle{Attention is all you need}.
\newblock \bibinfo{journal}{\emph{Advances in neural information processing systems}}  \bibinfo{volume}{30} (\bibinfo{year}{2017}).
\newblock


\bibitem[Wang et~al\mbox{.}(2018a)]%
        {wang2018additive}
\bibfield{author}{\bibinfo{person}{Feng Wang}, \bibinfo{person}{Jian Cheng}, \bibinfo{person}{Weiyang Liu}, {and} \bibinfo{person}{Haijun Liu}.} \bibinfo{year}{2018}\natexlab{a}.
\newblock \showarticletitle{Additive margin softmax for face verification}.
\newblock \bibinfo{journal}{\emph{IEEE Signal Processing Letters}} \bibinfo{volume}{25}, \bibinfo{number}{7} (\bibinfo{year}{2018}), \bibinfo{pages}{926--930}.
\newblock


\bibitem[Wang et~al\mbox{.}(2018b)]%
        {wang2018cosface}
\bibfield{author}{\bibinfo{person}{Hao Wang}, \bibinfo{person}{Yitong Wang}, \bibinfo{person}{Zheng Zhou}, \bibinfo{person}{Xing Ji}, \bibinfo{person}{Dihong Gong}, \bibinfo{person}{Jingchao Zhou}, \bibinfo{person}{Zhifeng Li}, {and} \bibinfo{person}{Wei Liu}.} \bibinfo{year}{2018}\natexlab{b}.
\newblock \showarticletitle{Cosface: Large margin cosine loss for deep face recognition}. In \bibinfo{booktitle}{\emph{Proceedings of the IEEE Conference on Computer Vision and Pattern Recognition}}. \bibinfo{pages}{5265--5274}.
\newblock


\bibitem[Wei et~al\mbox{.}(2020)]%
        {wei2020maggan}
\bibfield{author}{\bibinfo{person}{Yi Wei}, \bibinfo{person}{Zhe Gan}, \bibinfo{person}{Wenbo Li}, \bibinfo{person}{Siwei Lyu}, \bibinfo{person}{Ming-Ching Chang}, \bibinfo{person}{Lei Zhang}, \bibinfo{person}{Jianfeng Gao}, {and} \bibinfo{person}{Pengchuan Zhang}.} \bibinfo{year}{2020}\natexlab{}.
\newblock \showarticletitle{Maggan: High-resolution face attribute editing with mask-guided generative adversarial network}. In \bibinfo{booktitle}{\emph{Proceedings of the Asian Conference on Computer Vision}}.
\newblock


\bibitem[Wu et~al\mbox{.}(2021)]%
        {wu2021stylespace}
\bibfield{author}{\bibinfo{person}{Zongze Wu}, \bibinfo{person}{Dani Lischinski}, {and} \bibinfo{person}{Eli Shechtman}.} \bibinfo{year}{2021}\natexlab{}.
\newblock \showarticletitle{Stylespace analysis: Disentangled controls for stylegan image generation}. In \bibinfo{booktitle}{\emph{Proceedings of the IEEE/CVF Conference on Computer Vision and Pattern Recognition}}. \bibinfo{pages}{12863--12872}.
\newblock


\bibitem[Xia et~al\mbox{.}(2021)]%
        {xia2021tedigan}
\bibfield{author}{\bibinfo{person}{Weihao Xia}, \bibinfo{person}{Yujiu Yang}, \bibinfo{person}{Jing-Hao Xue}, {and} \bibinfo{person}{Baoyuan Wu}.} \bibinfo{year}{2021}\natexlab{}.
\newblock \showarticletitle{Tedigan: Text-guided diverse face image generation and manipulation}. In \bibinfo{booktitle}{\emph{Proceedings of the IEEE/CVF conference on computer vision and pattern recognition}}. \bibinfo{pages}{2256--2265}.
\newblock


\bibitem[Xiao et~al\mbox{.}(2018)]%
        {xiao2018elegant}
\bibfield{author}{\bibinfo{person}{Taihong Xiao}, \bibinfo{person}{Jiapeng Hong}, {and} \bibinfo{person}{Jinwen Ma}.} \bibinfo{year}{2018}\natexlab{}.
\newblock \showarticletitle{Elegant: Exchanging latent encodings with gan for transferring multiple face attributes}. In \bibinfo{booktitle}{\emph{Proceedings of the European conference on computer vision (ECCV)}}. \bibinfo{pages}{168--184}.
\newblock


\bibitem[Xu et~al\mbox{.}(2019)]%
        {xu2019generating}
\bibfield{author}{\bibinfo{person}{Caie Xu}, \bibinfo{person}{Ying Tang}, \bibinfo{person}{Masahiro Toyoura}, \bibinfo{person}{Jiayi Xu}, {and} \bibinfo{person}{Xiaoyang Mao}.} \bibinfo{year}{2019}\natexlab{}.
\newblock \showarticletitle{Generating Users’ Desired Face Image Using the Conditional Generative Adversarial Network and Relevance Feedback}.
\newblock \bibinfo{journal}{\emph{IEEE Access}}  \bibinfo{volume}{7} (\bibinfo{year}{2019}), \bibinfo{pages}{181458--181468}.
\newblock


\bibitem[Yang et~al\mbox{.}(2021b)]%
        {yang2021l2m}
\bibfield{author}{\bibinfo{person}{Guoxing Yang}, \bibinfo{person}{Nanyi Fei}, \bibinfo{person}{Mingyu Ding}, \bibinfo{person}{Guangzhen Liu}, \bibinfo{person}{Zhiwu Lu}, {and} \bibinfo{person}{Tao Xiang}.} \bibinfo{year}{2021}\natexlab{b}.
\newblock \showarticletitle{L2m-gan: Learning to manipulate latent space semantics for facial attribute editing}. In \bibinfo{booktitle}{\emph{Proceedings of the IEEE/CVF Conference on Computer Vision and Pattern Recognition}}. \bibinfo{pages}{2951--2960}.
\newblock


\bibitem[Yang et~al\mbox{.}(2021a)]%
        {yang2021discovering}
\bibfield{author}{\bibinfo{person}{Huiting Yang}, \bibinfo{person}{Liangyu Chai}, \bibinfo{person}{Qiang Wen}, \bibinfo{person}{Shuang Zhao}, \bibinfo{person}{Zixun Sun}, {and} \bibinfo{person}{Shengfeng He}.} \bibinfo{year}{2021}\natexlab{a}.
\newblock \showarticletitle{Discovering interpretable latent space directions of gans beyond binary attributes}. In \bibinfo{booktitle}{\emph{Proceedings of the IEEE/CVF conference on computer vision and pattern recognition}}. \bibinfo{pages}{12177--12185}.
\newblock


\bibitem[Yao et~al\mbox{.}(2021)]%
        {yao2021latent}
\bibfield{author}{\bibinfo{person}{Xu Yao}, \bibinfo{person}{Alasdair Newson}, \bibinfo{person}{Yann Gousseau}, {and} \bibinfo{person}{Pierre Hellier}.} \bibinfo{year}{2021}\natexlab{}.
\newblock \showarticletitle{A latent transformer for disentangled face editing in images and videos}. In \bibinfo{booktitle}{\emph{Proceedings of the IEEE/CVF international conference on computer vision}}. \bibinfo{pages}{13789--13798}.
\newblock


\bibitem[Zaltron et~al\mbox{.}(2020)]%
        {zaltron2020cg}
\bibfield{author}{\bibinfo{person}{Nicola Zaltron}, \bibinfo{person}{Luisa Zurlo}, {and} \bibinfo{person}{Sebastian Risi}.} \bibinfo{year}{2020}\natexlab{}.
\newblock \showarticletitle{Cg-gan: An interactive evolutionary gan-based approach for facial composite generation}. In \bibinfo{booktitle}{\emph{Proceedings of the AAAI Conference on Artificial Intelligence}}, Vol.~\bibinfo{volume}{34}. \bibinfo{pages}{2544--2551}.
\newblock


\bibitem[Zhang et~al\mbox{.}(2018)]%
        {zhang2018generative}
\bibfield{author}{\bibinfo{person}{Gang Zhang}, \bibinfo{person}{Meina Kan}, \bibinfo{person}{Shiguang Shan}, {and} \bibinfo{person}{Xilin Chen}.} \bibinfo{year}{2018}\natexlab{}.
\newblock \showarticletitle{Generative adversarial network with spatial attention for face attribute editing}. In \bibinfo{booktitle}{\emph{Proceedings of the European conference on computer vision (ECCV)}}. \bibinfo{pages}{417--432}.
\newblock


\bibitem[Zheng et~al\mbox{.}(2020)]%
        {zheng2020decoding}
\bibfield{author}{\bibinfo{person}{Xiao Zheng}, \bibinfo{person}{Wanzhong Chen}, \bibinfo{person}{Mingyang Li}, \bibinfo{person}{Tao Zhang}, \bibinfo{person}{Yang You}, {and} \bibinfo{person}{Yun Jiang}.} \bibinfo{year}{2020}\natexlab{}.
\newblock \showarticletitle{Decoding human brain activity with deep learning}.
\newblock \bibinfo{journal}{\emph{Biomedical Signal Processing and Control}}  \bibinfo{volume}{56} (\bibinfo{year}{2020}), \bibinfo{pages}{101730}.
\newblock


\end{thebibliography}

\end{document}